%% file: ms.tex
\documentclass[sigconf]{acmart}
\pdfoutput=1
\renewcommand\footnotetextcopyrightpermission[1]{} 
\usepackage{booktabs} 
\usepackage{multirow}
\usepackage{multicol}
\usepackage{comment}
\usepackage{subfigure}
\usepackage{balance}

\setcopyright{none}

\begin{document}
\fancyhead{} 
\settopmatter{printacmref=false}

\title{Semantic Human Matting}

\author{Quan Chen$^1$, Tiezheng Ge$^1$, Yanyu Xu$^{1,2}$, Zhiqiang Zhang$^1$, Xinxin Yang$^1$, Kun Gai$^1$}
\affiliation{%
  \institution{$^1$Alibaba Group, Beijing, China; $^2$ShanghaiTech University, Shanghai, China}
}
\email{{chenquan.cq, \space tiezheng.gtz, zhang.zhiqiang, xinxin.yxx}@alibaba-inc.com}
\email{xuyy2@shanghaitech.edu.cn,jingshi.gk@taobao.com}


\begin{abstract}
    Human matting, high quality extraction of humans from natural images, is crucial for a wide variety of applications.
	Since the matting problem is severely under-constrained, most previous methods require user interactions to take user designated trimaps or scribbles as constraints.
	This user-in-the-loop nature makes them difficult to be applied to large scale data or time-sensitive scenarios.
	In this paper, instead of using explicit user input constraints, we employ implicit semantic constraints learned from data and propose an automatic human matting algorithm \emph{Semantic Human Matting} (SHM).
	SHM is the first algorithm that learns to jointly fit both semantic information and high quality details with deep networks.
	In practice, simultaneously learning both coarse semantics and fine details is challenging.
	We propose a novel fusion strategy which naturally gives a probabilistic estimation of the alpha matte.
	We also construct a very large dataset with high quality annotations consisting of 35,513 unique foregrounds to facilitate the learning and evaluation of human matting.
	Extensive experiments on this dataset and plenty of real images show that SHM achieves comparable results with state-of-the-art interactive matting methods.
\end{abstract}

%
%




\keywords{Matting, Human Matting, Semantic Segmentation}

\maketitle

\input{semantic-human-matting-body-conf}

\balance
\input{ms.bbl}


\end{document}

%% file: semantic-human-matting-body-conf.tex
\section{Introduction}

\begin{figure}
  \centering
  \includegraphics[width=0.99\linewidth]{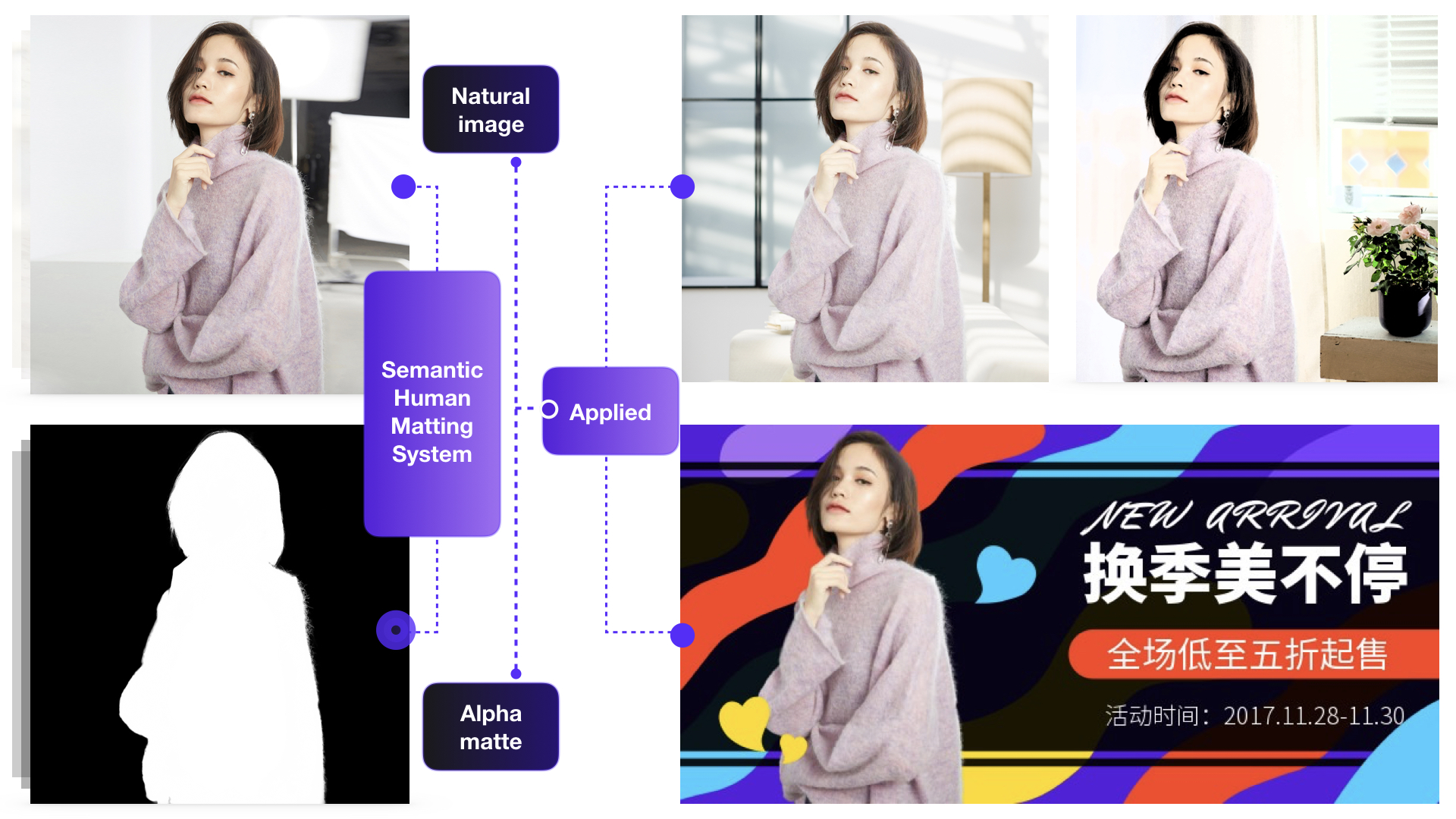}\\
  \caption{Semantic Human Matting (SHM) and its applications. SHM takes natural image (top left) as input and outputs corresponding alpha matte (bottom left). The predicted alpha matte could be applied to background editting (top right) and smart creative composition (bottom right)}
  \label{human_matting}
\end{figure}

Human matting, which aims at extracting humans from natural images with high quality, has a wide variety of applications, such as mixed reality, smart creative composition, live streaming, film production, \emph{etc}.
For example, in an e-commerce website, smart creative composition provides personalized creative image to customers.
This requires extracting fashion models from huge amount of original images and re-compositing them with new creative designes.
In such a scenario, due to the huge volume of images to be processed and in pursuit of a better customer experience,
it is critical to have an automatic high quality extraction method.
Fig.~\ref{human_matting} gives an example of smart creative composition with automatic human matting in a real-world e-commerce website.

Designing such an automatic method is not a trivial task.
One may think of turning to either semantic segmentation or image matting techniques.
However, neither of them can be used by itself to reach a satisfactory solution.
On the one hand, semantic segmentation, which directly identifies the object category of each pixel, 
usually focuses on the coarse semantics and is prone to blurring structural details.
On the other hand, image matting, widely adopted for fine detail extractions, usually requires user interactions and therefore is not suitable in data-intensive or time-sensitive scenarios such as smart creative composition.
More specifically,
for an input image $I$, matting is formulated as a decomposition into foreground $F$, background $B$ and alpha matte $\alpha$ with a linear blend assumption:
\begin{equation}
\label{equa_composition}
I = \alpha F + (1-\alpha) B , \quad \alpha \in [0, 1]
\end{equation}
where for color images, there are 7 unknown variables but only 3 known variables, and thus, this decomposition is severely under-constrained~\cite{levin2008closed}.
Therefore, most matting algorithms~\cite{levin2008closed,chen2013knn,aksoy2017designing,xu2017deep} need to take user designated trimaps or scribbles as extra constraints.

In this paper, we propose a unified method, Semantic Human Matting (SHM), which integrates a semantic segmentation module with a deep learning based matting module to automatically extract the alpha matte of humans.
The learned semantic information distinguishing foreground and background is employed as an implicit constraint
to a deep matting network which complements the capability of detail extraction.
A straightforward way to implement such a method is to train these two modules separately and feed the segmentation results as trimaps into the matting network.
However, this intuitive approach does not work well~\cite{shen2016deep}.
The reason is that 
semantic segmentation aims at classifying each pixel and is able to roughly distinct humans from background, whereas the goal of matting is to assign to each pixel a more fine grained float opacity value of foreground without determining the semantics.
They are responsible for recovering coarse segmentations and fine details respectively, and therefore they need to be carefully handled in order to cooperate properly towards high quality human matting.
Shen \emph{et al.}~\cite{shen2016deep} use a closed form matting~\cite{levin2008closed} layer through which the semantic information can directly propagate and constitute the final result.
But with deep learning based matting, the matting module is highly nonlinear and trained to focus on structural patterns of details, thus the semantic information from input hardly retains.
To combine the coarse semantics and fine matting details exquisitely, we propose a novel fusion strategy which naturally gives a probabilistic estimation of the alpha matte.
It can be viewed as an adaptive ensemble of both high and low level results on each pixel.
Further, with this strategy, the whole network automatically amounts the final training error to the coarse and the fine, and thus can be trained in an end-to-end fashion.

We also constructed a very large dataset with high quality annotations for the human matting task.
Since annotating details is difficult and time-consuming, high quality datasets for human matting are valuable and scarce.
The most popular alphamatting.com dataset~\cite{rhemann2009perceptually} has made significant contributions to the matting research. Unfortunately it only consists of 27 training images and 8 testing images.
Shen \emph{et al}.~\cite{shen2016deep} created a dataset of 2,000 images, but it only contains portrait images.
Besides, the groundtruth images of this dataset are generated with closed form matting~\cite{rhemann2009perceptually} and KNN matting~\cite{chen2013knn} and therefore can be potentially biased.
Recently, Xu \emph{et al}.~\cite{xu2017deep} built a large high quality matting dataset, with 202 distinct human foregrounds.
To increase the volume and diversity of human matting data that benefit the learning and evaluation of human matting, we collected another 35,311 distinct human images with fine matte annotations.
All human foregrounds are composited with different backgrounds and the final dataset includes 52,511 images for training and 1,400 images for testing.
More details of this dataset will be discussed in Section~\ref{dataset}.

Extensive experiments are conducted on this dataset to empirically evaluate the effectiveness of our method.
Under the commonly used metrics of matting performance, our method can achieve comparable results with the state-of-the-art interactive matting methods~\cite{levin2008closed,chen2013knn,aksoy2017designing,xu2017deep}.
Moreover, we demonstrate that our learned model generalizes to real images with justifying plenty of natural human images crawled from the Internet.

To summarize, the main contributions of our work are three-fold:

1. To the best of our knowledge, SHM is the first automatic matting algorithm that learns to jointly fit both semantic information and high quality details with deep networks.
Empirical studies show that SHM achieves comparable results with the state-of-the-art interactive matting methods.

2. A novel fusion strategy, which naturally gives a probabilistic estimation of the alpha matte, is proposed to make the entire network properly cooperate.
It adaptively ensembles coarse semantic and fine detail results on each pixel which is crucial to enable end-to-end training.

3. A large scale high quality human matting dataset is created.
It contains 35,513 unique human images with corresponding alpha mattes.
The dataset not only enables effective training of the deep network in SHM but also contributes with its volume and diversity to the human matting research.

\section{Related works}

In this section, we will review semantic segmentation and image matting methods that most related to our works.

Since Long \emph{et al}. \cite{long2015fully} use Fully Convolutional Network (FCN) to predict pixel level label densely and improve the segmentation accuracy by a large margin, FCN has became the main framework for semantic segmentation and kinds of techniques have been proposed by researchers to improve the performance. Yu \emph{et al}.~\cite{yu2015multi} propose dilated convolutions to increase the receptive filed of the network without spatial resolution decrease, which is demonstrated effective for pixel level prediction. Chen \emph{et al}.~\cite{chen2016deeplab} add fully connected CRFs on the top of network as post processing to alleviate the "hole" phenomenon of FCN. In PSPNet~\cite{zhao2017pyramid}, network-in pyramid pooling module is proposed to acquire global contextual prior. Peng \emph{et al}.~\cite{peng2017large} state that using large convolutional kernels and boundary refinement block can improve the pixel level classification accuracy while maintaining precise localization capacity. With the above improvements, FCN based models trained on large scale segmentation datasets, such as VOC \cite{pascal-voc-2012} and COCO \cite{lin2014microsoft}, have achieved the top performances in semantic segmentation. However, these models can not be directly applied to semantic human matting for the following reasons.
1) The annotations of current segmentation datasets are relative "coarse" and "hard" to matting task. Models trained on these datasets do not satisfy the accuracy requirement of pixel level location and floating level alpha values for matting.
2) Pixel level classification accuracy is the only consideration during network architecture and loss design in semantic segmentation. This leads the model prone to blurring complex structural details which is crucial for matting performance.

In the past decades, researchers have developed variety of general matting methods for natural images. Most methods predict the alpha mattes through sampling \cite{chuang2001bayesian,wang2007optimized,gastal2010shared,he2011global,shahrian2013improving} or propagating \cite{sun2004poisson,grady2005random,levin2008closed,chen2013knn,aksoy2017designing} on color or low-level features. With the rise of deep learning in computer vision community, several CNN based methods \cite{cho2016natural,xu2017deep} have been proposed for general image matting. Cho \emph{et al}. \cite{cho2016natural} design a convolutional neural network to reconstruct the alpha matte by taking the results of the closed form matting \cite{levin2008closed} and KNN matting \cite{chen2013knn} along with the normalized RGB color image as inputs. Xu \emph{et al}. \cite{xu2017deep} directly predict the alpha matte with a pure encoder decoder network which takes the RGB image and trimap as inputs and achieve the state-of-the-art results.
However, all the above general image matting methods need scribbles or trimap obtained from user interactions as constraints and so they can not be applied in automatic way.

Recently, several works \cite{shen2016deep,zhu2017fast} have been proposed to make an automatic matting system.
Shen \emph{et al}. \cite{shen2016deep} use closed form matting~\cite{levin2008closed} with CNN to automatically obtain the alpha mattes of portrait images and back propagate the errors to the deep convolutional network. Zhu et al.~\cite{zhu2017fast} follow the similar pipeline while designing a smaller network and a fast filter similar to guided filter~\cite{he2010guided} for matting to deploy the model on mobile phones.
Despite our method and the above two works both use CNNs to learn semantic information instead of manual trimaps to automate the matting process, our method is quite different from theirs:

\begin{figure*}[hbt]
  \includegraphics[width=0.95\linewidth]{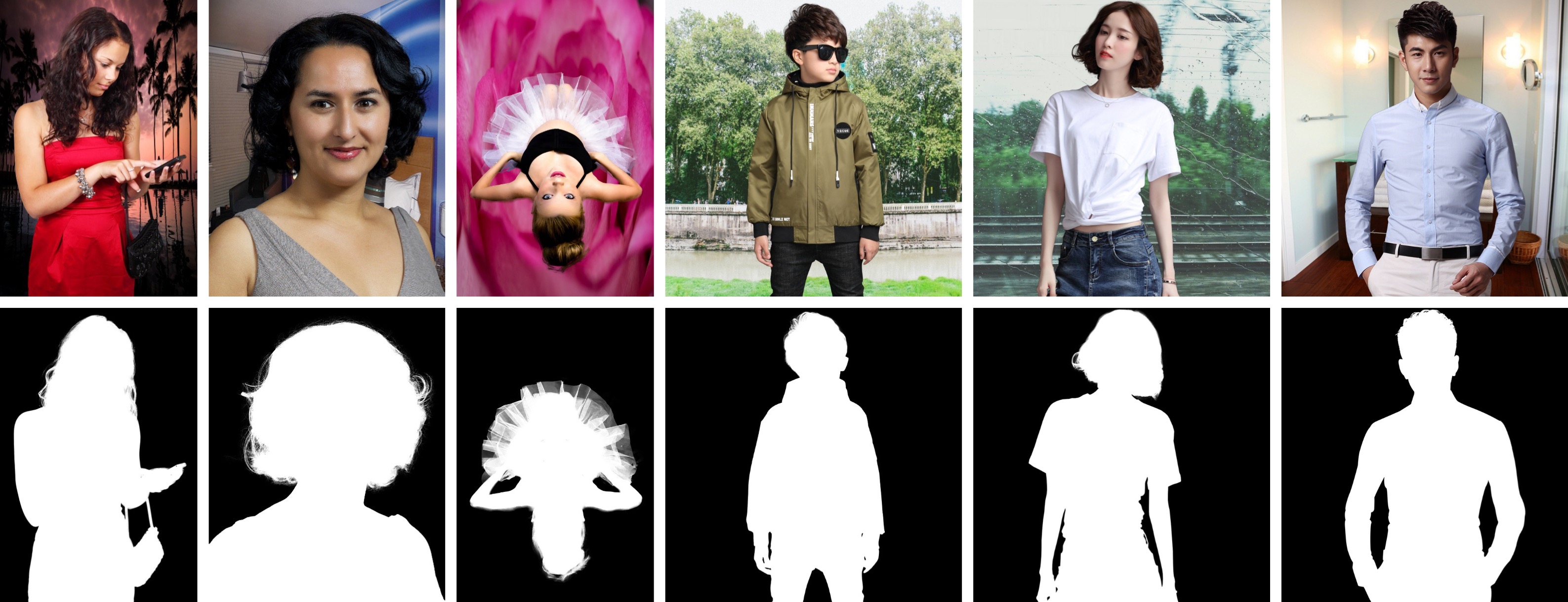}
  \caption{\label{dataset_intro} Compositional images and corresponding alpha mattes in our dataset. The first three columns come from the general matting dataset create by Xu \emph{et al}. \cite{xu2017deep} and the last three columns come from model images we collected from an e-commerce website. It's worth notting that images shown here are all resized to the same height.}
\end{figure*}

1) Both the above methods use the traditional methods as matting module, which compute the alpha matte by solving the matting equation (Eq.~\ref{equa_composition}) and may introduce artifacts when the distributions of foreground and background color overlap~\cite{xu2017deep}.
We employ a FCN as matting module so as to directly learn complex details in a wide context which have been shown much more robust by~\cite{xu2017deep}.
2) By solving the matting equation, these method can directly affect the final perdition with the input constraints and thus propagate back the errors.
However, when the deep matting network is adopted, the cooperation of coarse semantics and find details must be explicitly handled.
Thus a novel fusion strategy is proposed and enables the end-to-end training of the entire network.

\section{Human matting dataset}\label{dataset}

\begin{table}[t]
\caption{Configuration of our human matting dataset.}
\centering
\begin{tabular}{c cc cc}
\toprule
   \multirow{2}{*}{Data Source}  & \multicolumn{2}{c}{Train Set}  & \multicolumn{2}{c}{Test Set} \\
             &   $\#$Foreground         & $\#$Image  &   $\#$Foreground    & $\#$Image \\ \midrule
   DIM\cite{xu2017deep} &  182        & 18,200      &   20    & 400   \\
   Model                &  34,311      & 34,311      &   1,000  & 1,000  \\
   Total                &  34,493      & 52,511      &   1,020  & 1,400  \\ \bottomrule

\end{tabular}
\label{tab:dataset_config}
\end{table}


As a newly defined task in this paper, the first challenge is that semantic human matting encounters the lack of data.
To address it, we create a large scale high quality human matting dataset.
The foregrounds in this dataset are humans with some accessories(\emph{e.g.}, cellphones, handbags).
And each foreground is associated with a carefully annotated alpha matte.
Following Xu \emph{et al}.\cite{xu2017deep}, foregrounds are composited onto new backgrounds to create a human matting dataset with 52,511 images in total.
Some sample images in our dataset are shown in Fig. \ref{dataset_intro}.

In details, the foregrounds and corresponding alpha matte images in our dataset comprise:
\begin{itemize}

\item \textbf{Fashion Model dataset.}
 More than 188k fashion model images are collected from an e-commerce website, whose alpha mattes are annotated by sellers in accordance with
 commercial quality. Volunteers are recruited to carefully inspect and double-check the mattes to remove those even only with small flaws. It takes almost 1,200 hours to select 35,311 images out of them. The low pass rate (18.88 \%) guarantees the high standard of the alpha matte in our dataset.

\item \textbf{Deep Image Matting (DIM) dataset~\cite{xu2017deep}.} We also select all the images that only contain human from DIM dataset, resulting 202 foregrounds.
\end{itemize}
The background images are  from COCO dataset and the Internet.
We ensure that background images do not contain humans.
The foregrounds are split into train/test set, and the configuration is shown in Table \ref{tab:dataset_config}.
Following \cite{xu2017deep}, each foreground is composited with N backgrounds.
For foregrounds from Fashion Model dataset, due to their large number, N is set to 1 for both training and testing dataset.
For foregrounds from DIM dataset, N is set to 100 for training dataset and 20 for testing dataset, as in \cite{xu2017deep}.
All the background images are randomly selected and unique.

\newcommand{\tabincell}[2]{\begin{tabular}{@{}#1@{}}#2\end{tabular}}
\begin{table}[t]
\caption{The properties of the existing matting datasets.}
\centering
\begin{tabular}{cccc}
  \toprule
  Datasets & Foreground  & $\#$Image & Annotation \\
  \midrule
  alpha matting~\cite{rhemann2009perceptually}    &    35 Objects       &    35   &  Manually \\  
  Shen et al.~\cite{shen2016deep} &  2,000 Portraits       &   2,000    & CF \cite{levin2008closed}, KNN \cite{chen2013knn} \\
  DIM~\cite{xu2017deep}  &  493 Objects  &   49,300  & Manually  \\
  Our dataset &  35,513 Humans      & 52,511        & Manually  \\ 
  \bottomrule
\end{tabular}
\label{tab:dataset_components}
\end{table}

\begin{figure*}[hbt]
  \includegraphics[width=0.95\linewidth]{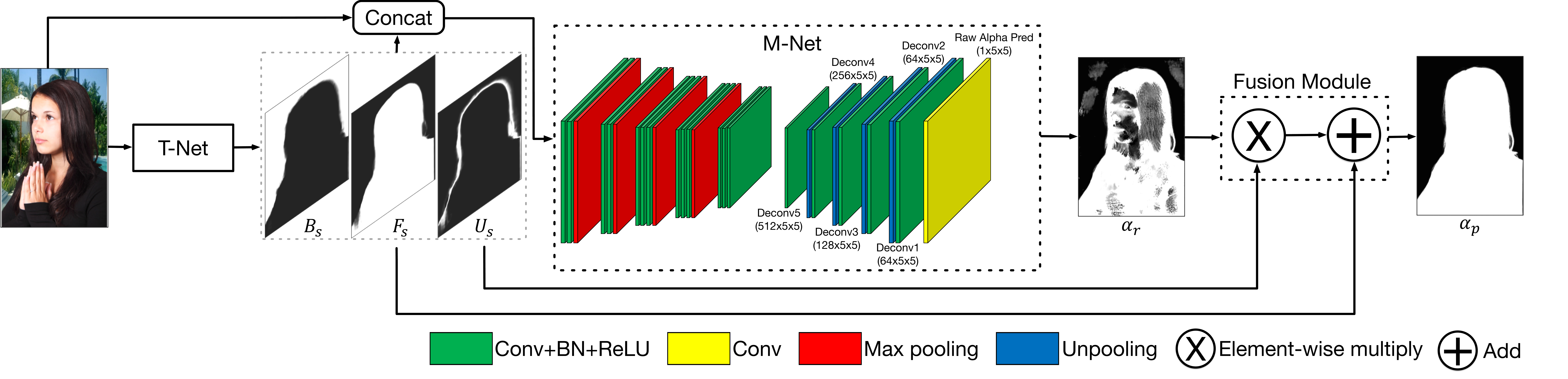}
  \caption{Overview of our semantic human matting method.
  	Given an input image, a \emph{T-Net}, which is implemented as PSPNet-50, is used to predict the 3-channel trimap. The predicted trimap is then concatenated with the original image and fed into the \emph{M-Net} to predict the raw alpha matte.
  	Finally, both the predicted trimap and raw alpha matte are fed into the \emph{Fusion Module} to generate the final alpha matte according to Eq. \ref{equa_t_m_fusion}.
  	The entire network is trained in an end-to-end fashion.
  }
  \label{pipeline}
\end{figure*}

Table \ref{tab:dataset_components} shows the comparisons of basic properties between existing matting datasets and ours. Compared with previous matting datasets, our dataset differs in the following aspects:
1) The existing matting datasets contain hundreds of foreground objects, while our dataset contains 35,513 different foregrounds which is much larger than others;
2) In order to deal with the human matting task, foregrounds containing human body are needed.
However, DIM\cite{xu2017deep} dataset only contains 202 human objects.
The dataset proposed by Shen et al. \cite{shen2016deep} consists of portraits, which are limited to heads and part of shoulders.
In contrast, our dataset has a larger diversity that might cover the whole human body, \emph{i.e.} head, arms, legs \emph{etc}. in various poses, which is essential for human matting;
3) Unlike the dataset of Shen et al \cite{shen2016deep} which is annotated by Closed From \cite{levin2008closed}, KNN \cite{chen2013knn} and therefore can be potentially biased, all 35,513 foreground objects are manually annotated and carefully inspected, which guarantees the high quality alpha mattes and ensures the semantic integrity and unicity.
The dataset not only enables effective training of the deep network in SHM but also contributes with its volume and diversity to the human matting research.

\section{Our method}

Our SHM is targeted to automatically pull the alpha matte of a specific semantic pattern---the humans.
Fig~\ref{pipeline} shows its pipeline.
The SHM takes an image (usually 3 channels representing RGB) as input, and directly outputs a 1-channel alpha matte image with identical size of input.
Note that no auxiliary information (\emph{e.g.} trimap and scribbles) is required.

SHM aims to simultaneously capture both coarse semantic classification information and fine matting details.
We design two subnetworks to separately handle these two tasks.
The first one, named \emph{T-Net}, is responsible to do pixel-wise classification among foreground, background and unknown regions; while the second one, named \emph{M-Net}, takes in the output of \emph{T-Net} as semantic hint and describes the details by generating the raw alpha matte image.
The outputs of \emph{T-Net} and \emph{M-Net} are fused by a novel \emph{Fusion Module} to generate the final alpha matte result.
The whole networks are trained jointly in an end-to-end manner.
We describe these submodules in detail in the following subsections.

\subsection{Trimap generation: T-Net}

The \emph{T-Net} plays the role of semantic segmentation in our task and roughly extract foreground region.
Specifically, we follow the traditional trimap concept and define a 3-class segmentation---the foreground, background and unknown region.
Therefore, the output of \emph{T-Net} is a 3-channel map indicating the possibility that each pixel belongs to each of the 3 classes.
In general, \emph{T-Net} can be implemented as any of the state-of-the-art semantic segmentation networks~\cite{long2015fully,yu2015multi,chen2016deeplab,zhao2017pyramid,peng2017large}.
In this paper, we choose PSPNet-50~\cite{zhao2017pyramid} for its efficacy and efficiency.

\subsection{Matting network: M-Net}


Similar to general matting task~\cite{xu2017deep}, the \emph{M-Net} aims to capture detail information and generate alpha matte.
The \emph{M-Net} takes the concatenation of 3-channel images and the 3-channel segmentation results from \emph{T-Net} as 6-channel input.
Note that it differs from DIM~\cite{xu2017deep} which uses 3-channel images plus 1-channel trimap (with 1, 0.5, 0 to indicate foreground, unknown region and background respectively) as 4-channel input.
We use 6-channel input since it conveniently fits the output of \emph{T-Net} and we empirically find that with 6-channel or 4-channel input have nearly equal performance.




As shown in Fig.~\ref{pipeline}, the \emph{M-Net} is a deep convolutional encoder-decoder network. The encoder network has 13 convolutional layers and 4 max-pooling layers, while the decoder network has 6 convolutional layers and 4 unpooling layers. The hyper-parameters of encoder network are the same as the convolutional layers of VGG16 classification network expect for the "conv1" layer in VGG16 that has 3 input channels whereas 6 in our M-Net.
The structure of \emph{M-Net} differs from DIM~\cite{xu2017deep} in following aspects:
1) \emph{M-Net} has 6-channel instead of 4-channel inputs; 2) Batch Normalization is added after each convolutional layer to accelerate convergence; 3) "conv6" and "deconv6" layers are removed since these layers have large number of parameters and are prone to overfitting.

\subsection{Fusion Module}
The deep matting network takes the predicted trimap as input and directly computes the alpha matte.
However, as shown in Fig.~\ref{pipeline}, it focuses on the unknown regions and recovers structual and textural details only.
The semantic information of foreground and background is not retained well.
In this section, we describe the fusion strategy in detail.

We use $F$, $B$ and $U$ to denote the foreground, background and unknown region channel that predicted by T-Net before softmax. Thus the probability map of foreground $F_s$ can be written as
\begin{equation}
\label{equa_foreground_score}
F_{s} = \frac{exp(F)}{exp(F) +exp(B) + exp(U)}
\end{equation}
We can obtain $B_s$ and $U_s$ in the similarly way.
It is obvious that $F_s + B_s + U_s = \textbf{1}$, where \textbf{1} denotes an all-1 matrix that has the same width and height of input image.
We use $\alpha_{r}$ to denote the output of M-Net.

Noting that the predicted trimap gives the probability distribution of each pixel belonging the three categories, foreground, background and unknown region.
When a pixel locates in the unknown region, which means that it is near the contour of a human and constitutes the complex structural details like hair, matting is required to accurately pull the alpha matte.
At this moment, we would like to use the result of matting network, $\alpha_r$, as an accurate prediction.
Otherwise, if a pixel locates outside the unknown region, then the conditional probability of the pixel belonging to the foreground is an appropriate estimation of the matte, \emph{i.e.}, $\frac{F_{s}}{F_{s} + B_{s}}$.
Considering that $U_s$ is the probability of each pixel belonging to the unknown region, a probabilistic estimation of alpha matte for all pixels can be written as
\begin{equation}
\label{equa_t_m_fusion_derivation}
\alpha_{p} = (\textbf{1} - U_{s})\frac{F_{s}}{F_{s} + B_{s}} + U_{s}\alpha_{r}
\end{equation}
where $\alpha_{p}$ denotes the output of \emph{Fusion Module}. As $F_{s} + B_{s} = 1 - U_{s}$, we can rewrite Eq.~\ref{equa_t_m_fusion_derivation} as
\begin{equation}
\label{equa_t_m_fusion}
\alpha_{p} = F_{s} + U_{s}\alpha_{r}
\end{equation}

Intuitively, this formulation shows that the coarse semantic segmentation is refined by the matting result with details, and the refinement is controlled explicitly by the unknown region probability.
We can see that when $U_s$ is close to 1, $F_s$ is close to 0, so $\alpha_p$ is approximated by $\alpha_r$, and
when $U_s$ is close to 0, $\alpha_p$ is approximated by $F_s$.
Thus it naturally combines the coarse semantics and fine details.
Furthermore, training errors can be readily propagated through to corresponding components, enabling the end-to-end training of the entire network.

\subsection{Loss}
Following Xu \emph{et al}.~\cite{xu2017deep}, we adopt the alpha prediction loss and compositional loss.
The alpha prediction loss is defined as the absolute difference between the groundtruth alpha $\alpha_g$ and predicted alpha $\alpha_p$.
And the compositional loss is defined as the absolute difference between the groundtruth compositional image values $c_g$ and predicted compositional image values $c_p$.
The overall prediction loss for $\alpha_p$ at each pixel is
\begin{equation}
\mathcal{L}_p = \gamma ||\alpha_p - \alpha_g||_1 + (1 - \gamma)||c_p - c_g||_1
\end{equation}
where $\gamma$ is set to 0.5 in our experiments.
It is worth noting that unlike Xu \emph{et al}.~\cite{xu2017deep} which only focus on unknown regions, in our automatic settings, the prediction loss is summed over the entire image.

In addition, we need to note that the loss $||F_{s} + U_{s}\alpha_{r} - \alpha_g||$ forms another decomposition problem of groundtruth matte, which is again under-constrained.
To get a stable solution to this problem, we introduce an extra constraint to keep the trimap meaningful.
A classification loss $\mathcal{L}_t$ for the trimap over each pixel is thus involved.

Finally, we get the total loss
\begin{equation}\label{total_loss}
\mathcal{L} = \mathcal{L}_p + \lambda \mathcal{L}_t
\end{equation}
where we just keep $\lambda$ to a small value to give a decomposition constraint, e.g. 0.01 throughout our paper.

\subsection{Implementation Detail}\label{sec_implementaton}

The pre-train technique~\cite{hinton2006fast} has been widely adopted in deep learning and shown its effectiveness.
We follow this common practice and first pre-train the two sub-netowrks \emph{T-Net} and \emph{M-Net} separately and then finetune the entire net in an end-to-end way.
Further, when pre-training the subnetwork, extra data with large amount specific to sub-tasks can also be empolyed to sufficiently train the models.
Note that the dataset used for pre-training should not overlap with the test set.

\paragraph{\textbf{T-Net pre-train}}
To train T-Net, we follow the common practice to generate the trimap ground truth by dilating the groundtruth alpha mattes.
In training phase, square patches are randomly cropped from input images and then uniformly resized to 400$\times$400.
To avoid overfitting, these samples are also augmented by randomly performing rotation and horizontal flipping.
As our \emph{T-Net} makes use of PSPNet50 which is based on ResNet50~\cite{he2016deep}, we initialize relevant layers with off-the-shelf model trained on ImageNet classification task and randomly initialize the rest layers.
The cross entropy loss for classification(\emph{i.e.}, $\mathcal{L}_t$ in Eq.~\ref{total_loss}) is employed.


\paragraph{\textbf{M-Net pre-train}} 
We follow deep matting network training pipeline as~\cite{xu2017deep} to pre-train \emph{M-Net}.
Again, the input of \emph{M-Net} is a 3-channel image with the 3-channel trimap generated by dilating and eroding the groundtruth alpha mattes.
Worth noting that, we find it is crucial for the performance of matting to augment the trimap with different kernel sizes of dilating and eroding, since it makes the result more robust to the various unknown region widths.
For data augmentation, the input images are randomly cropped and resized to 320$\times$320.
Entire DIM~\cite{xu2017deep} dataset is empployed during pre-training \emph{M-Net} regardless whether it contains humans, since the \emph{M-Net} focuses on the local pattern rather than the global semantic meaning.
The regression loss same as $\mathcal{L}_p$ term in Eq.~\ref{total_loss} is adopted.
   

\paragraph{\textbf{End-to-end training}}

End-to-end training is performed on human matting dataset and the model is initialized by pre-trained \emph{T-Net} and \emph{M-Net}.
In training stage, the input image is randomly cropped as 800$\times$800 patches and fed into \emph{T-Net} to obtain semantic predictions.
Considering that \emph{M-Net} needs to be more focused on details and trained with large diversity,
augmentations are performed on the fly to randomly crop different patches (320$\times$320, 480$\times$480, 640$\times$640 as in~\cite{xu2017deep}) and resize to 320$\times$320.
Horizontal flipping is also randomly adopted with 0.5 chance.
The total loss in Eq.~\ref{total_loss} is used.
For testing, the feed forward is conducted on the entire image without augmentation. More specifically, when the longer edge of the input image is larger than 1500 pixels, we first scale it to 1500 for the limitation of GPU memory. We then feed it to the network and finally rescale the predicted alpha matte to the size of the original input image for performance evaluation. In fact, we can alternatively perform testing on CPU for large images without losing resolution.

\section{Experiments}

\subsection{Experimental Setup}

We implement our method with PyTorch~\cite{paszke2017automatic} framework.
The \emph{T-Net} and \emph{M-Net} are first pre-trained and then fine-tuned end to end as described in Section~\ref{sec_implementaton}.
During end-to-end training phase,  we use Adam as the optimizer.
The learning rate is set to $10^{-5}$ and the batch size is 10.

\paragraph{\textbf{Dataset}}

We evaluate our method on the human matting dataset, which contains 52,511 training images and 1,400 testing images as described in Section~\ref{dataset}.

\paragraph{\textbf{Measurement}}

Four metrics are used to evaluate the quality of predicted alpha matte~\cite{rhemann2009perceptually}: SAD, MSE, Gradient error and Connectivity error.
SAD and MSE are obviously correlated to the training objective, and the Gradient  error and Connectivity error are proposed by~\cite{rhemann2009perceptually} to reflect perceptual visual quality by a human observer.
To be specific, we normalize both the predicted alpha matte and groundtruth to 0 to 1 when calculating all these metrics.
Further, all metrics are caculated over entire images instead of only within unknown regions and averaged by the pixel number.

\paragraph{\textbf{Baselines}}
In order to evaluate the effectiveness of our proposed methods, we compare our method with the following state-of-the-art matting methods\footnote{Implementations provided by their authors are used except for DIM. We implement the DIM network with the same structure as \emph{M-Net} except with 4 input channels for a fair comparision.}:
Closed Form (CF) matting \cite{levin2008closed} , KNN matting \cite{chen2013knn}, DCNN matting \cite{cho2016natural} , Information Flow Matting (IFM) \cite{aksoy2017designing} and Deep Image Matting (DIM) \cite{xu2017deep}.
Noting that, all these matting methods are interactive and need extra trimaps as input.
For a fair comparison, we provide them with predicted trimaps by the well pretrained \emph{T-Net}.
We denote these methods as \textbf{PSP50 + $X$}, where $X$ represents the above methods.

To demonstrate the results of applying semantic segmentation to matting problem, we also design the following baselines:
\begin{itemize}
  \item PSP50 Seg: a PSPNet-50 is used to extract humans via the predicted mask. The groundtruth mask used to train this network is obtained by binarizing the alpha matte with a threshold of 0.
  \item PSP50 Reg: a PSPNet-50 is trained to predict the alpha matte as regression with L1 loss.
\end{itemize}

\begin{table}
	\caption{The quantitative results on human matting testing dataset. The best results are emphasized in bold}
	\label{tab:quan_res_hmtds_no_trimap}
	\begin{tabular}{lllll}
		\toprule
		Methods & \tabincell{c}{SAD \\ ($\times10^{-3}$)} & \tabincell{c}{MSE \\ ($\times10^{-3}$)} & \tabincell{c}{Gradient \\ ($\times10^{-5}$)} & \tabincell{c}{Connectivity \\ ($\times 10^{-5}$)} \\
		\midrule
		PSP50 Seg & 14.821 & 11.530 & 52.336 & 44.854\\
		PSP50 Reg & 10.098 & 5.430 & 15.441 & 65.217\\
		PSP50+CF \cite{levin2008closed} & 8.809 & 5.218 & 21.819 & 43.927\\
		PSP50+KNN \cite{chen2013knn} & 7.806 & 4.390 & 20.476 & 56.328\\
		PSP50+DCNN \cite{cho2016natural} & 8.378 & 4.756 & 20.801 & 50.574\\
		PSP50+IFM \cite{aksoy2017designing} & 7.576 & 4.275 & 19.762 & 52.470\\
		PSP50+DIM \cite{xu2017deep} & 6.140 & 3.834 & 19.414 & 41.884\\
		\midrule
		Our Method & \textbf{3.833} & \textbf{1.534} & \textbf{5.179} & \textbf{36.513} \\
		\bottomrule
	\end{tabular}
\end{table}

\subsection{Performance Comparison}

In this section, we compare our method with the state-of-the-art matting methods with generated trimaps and designed baselines on the human matting testing dataset.
Trimaps are predicted by the pre-trained \emph{T-Net} and are provided to interactive matting methods.
The quantitative results are listed in Table~\ref{tab:quan_res_hmtds_no_trimap}.

The performances of binary segmentation and regression are poor.
Since complex structural details as well as the concepts of human are required in this task, the results show that it is hard to learn them simultaneously with a FCN network.
Using the trimaps predicted by the same PSP50 network, DIM outperforms the other methods, such as CF, KNN, DCNN and IFM. It is due to the strong capabilities of deep matting network to model complex context of images.
We can see that our method performs much better than all baselines.
The key reason is that our method successfully coordinate the coarse semantics and fine details with a probabilistic fusion strategy which enables a better end-to-end training.

\begin{figure*}[t]
	\footnotesize
	\begin{center}
		\begin{tabular}{cccccccc}
			\includegraphics[width=0.12\linewidth]{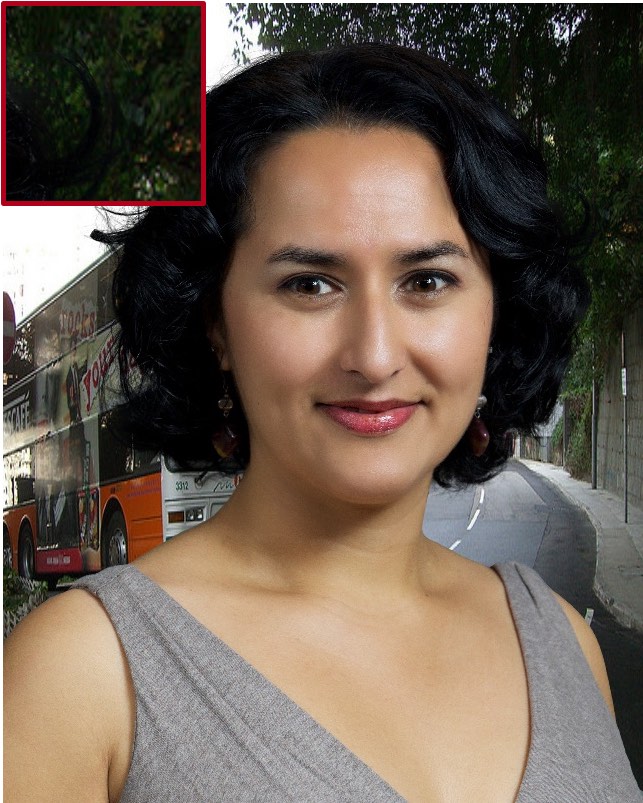} & \hspace{-0.3cm}
			\includegraphics[width=0.12\linewidth]{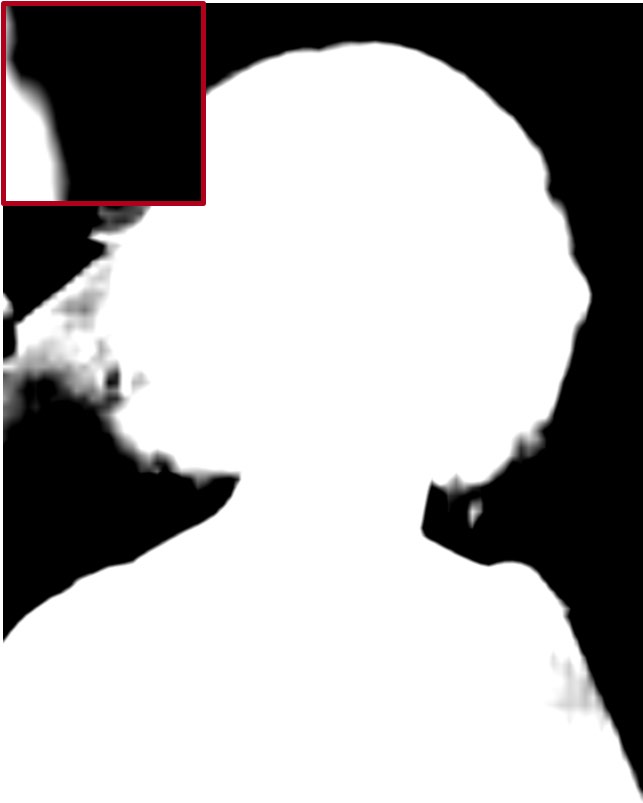} & \hspace{-0.3cm}
			\includegraphics[width=0.12\linewidth]{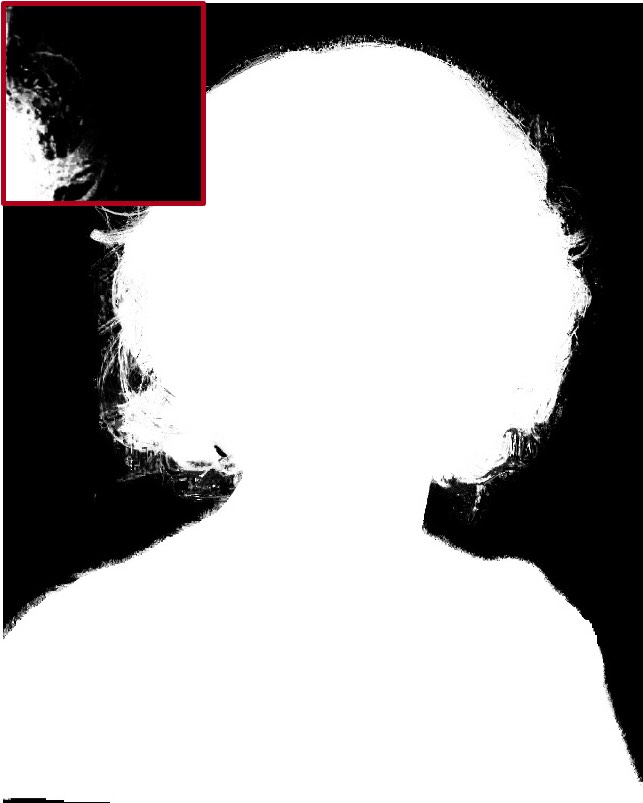} & \hspace{-0.3cm}
			\includegraphics[width=0.12\linewidth]{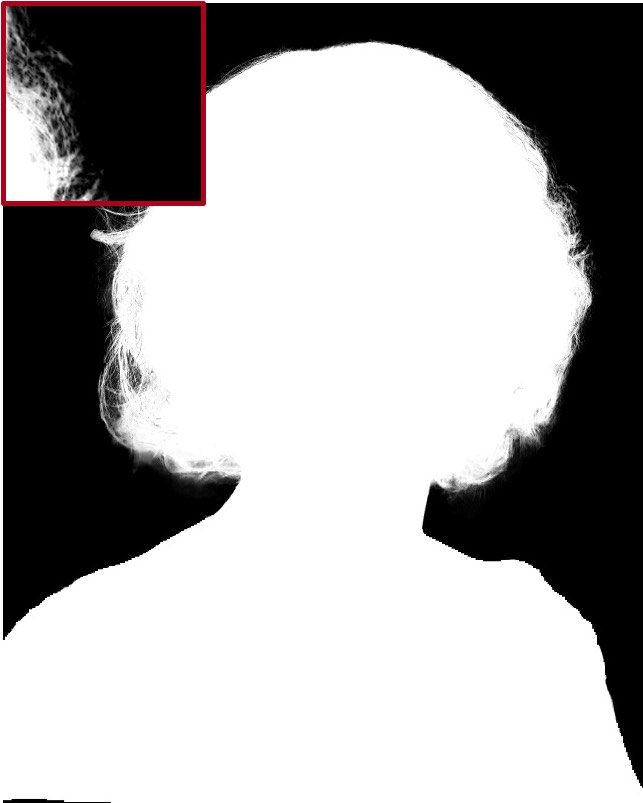} & \hspace{-0.3cm}
			\includegraphics[width=0.12\linewidth]{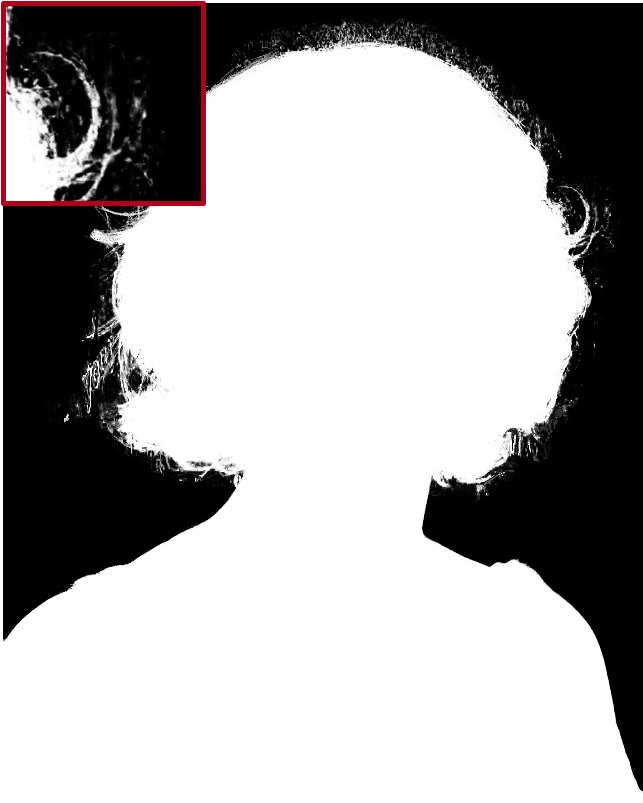} & \hspace{-0.3cm}
			\includegraphics[width=0.12\linewidth]{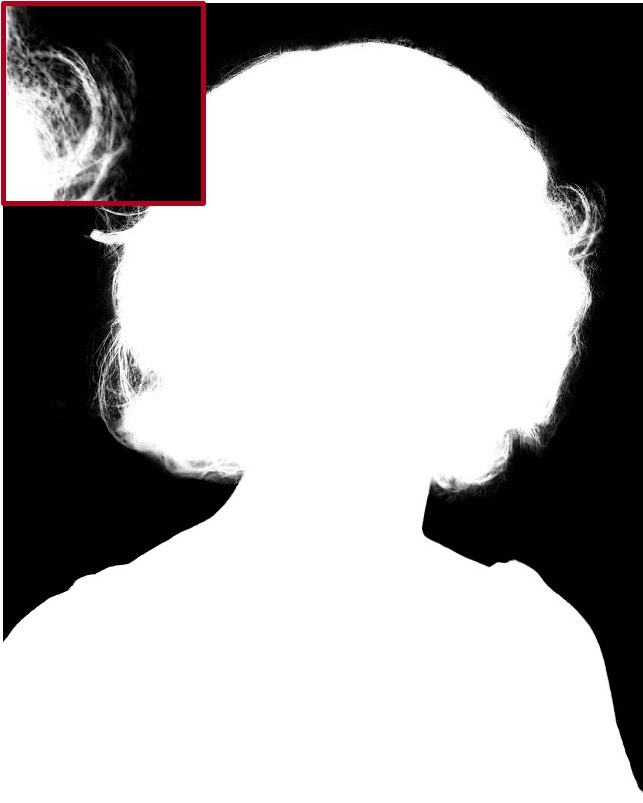} & \hspace{-0.3cm}
			\includegraphics[width=0.12\linewidth]{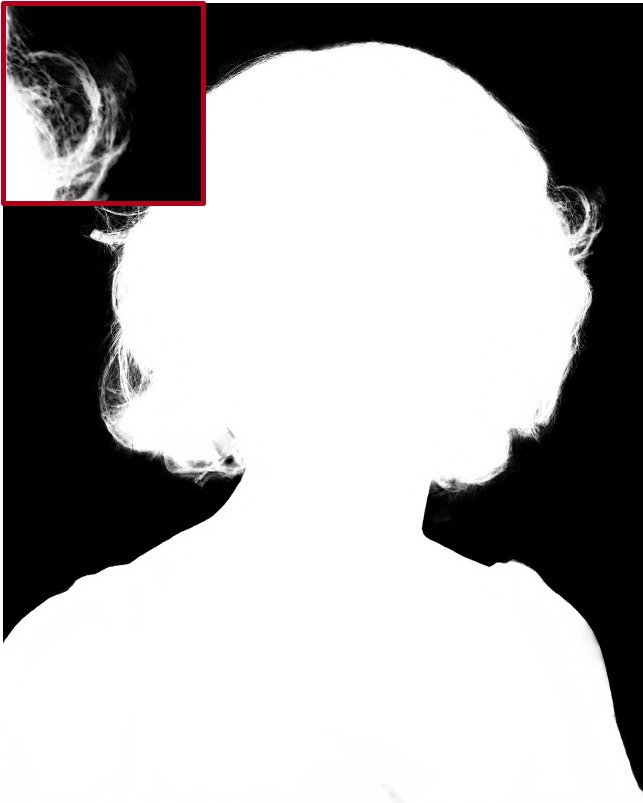} & \hspace{-0.3cm}
			\includegraphics[width=0.12\linewidth]{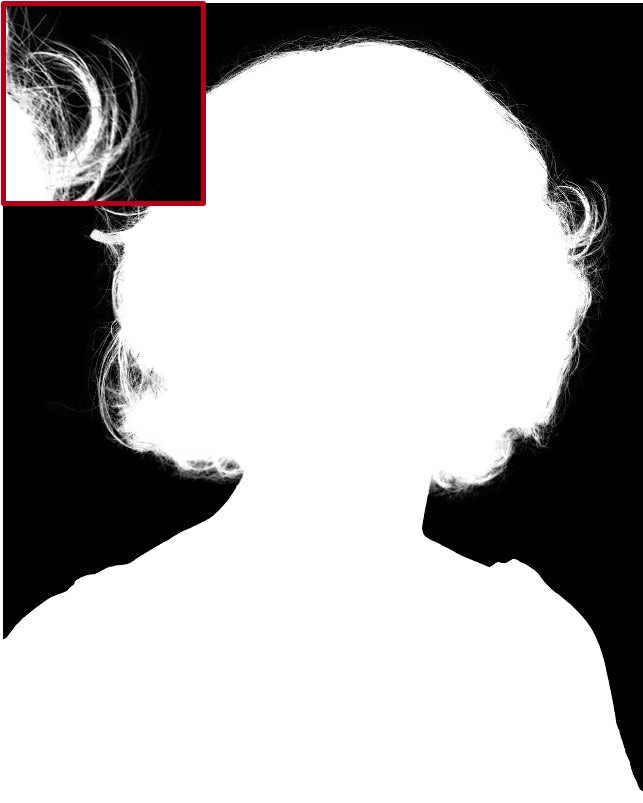}\\
			Image & PSP50 Reg & PSP50+IFM \cite{aksoy2017designing} & PSP50+DIM \cite{xu2017deep} & TrimapGT+IFM \cite{aksoy2017designing} & TrimapGT+DIM \cite{xu2017deep} & Our method & Alpha GT\\
			
			\includegraphics[width=0.12\linewidth]{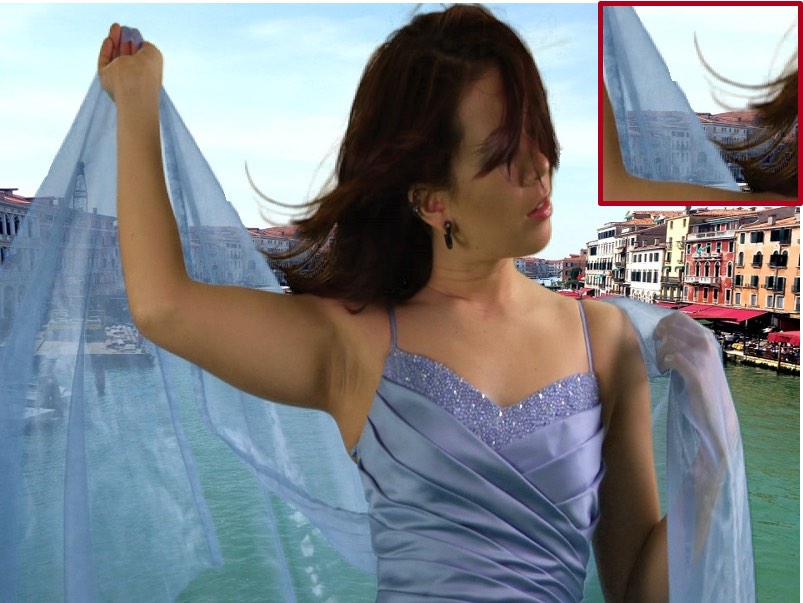} & \hspace{-0.3cm}
			\includegraphics[width=0.12\linewidth]{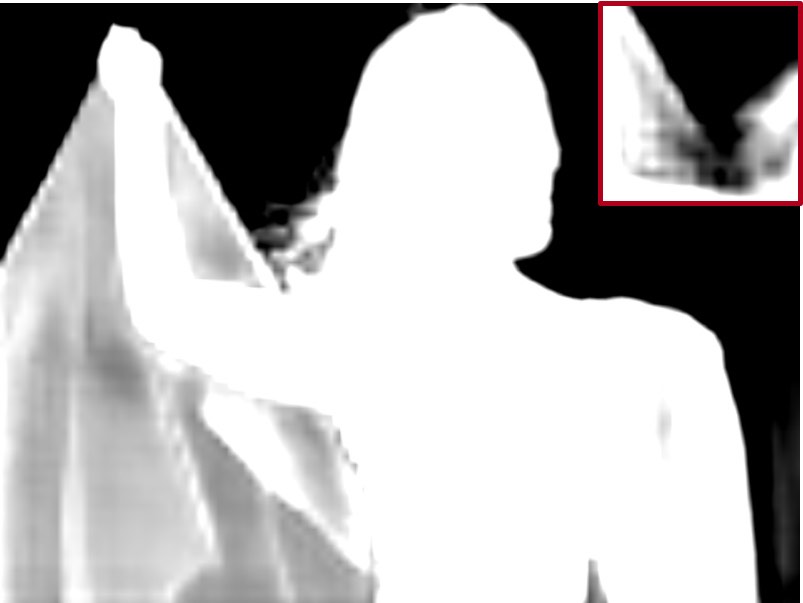} & \hspace{-0.3cm}
			\includegraphics[width=0.12\linewidth]{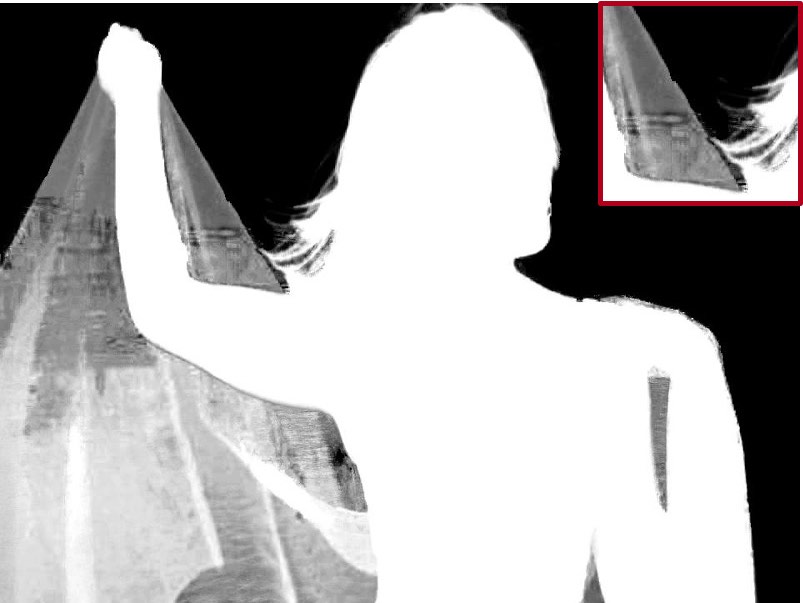} & \hspace{-0.3cm}
			\includegraphics[width=0.12\linewidth]{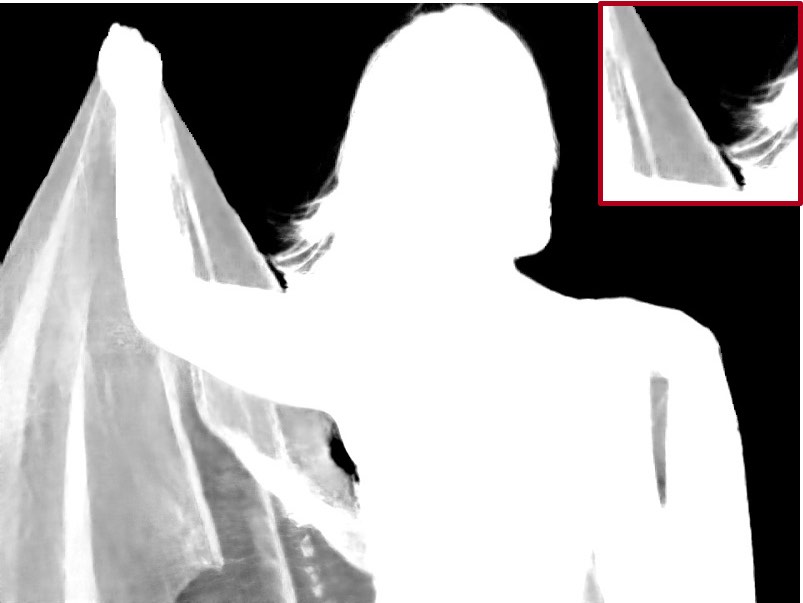} & \hspace{-0.3cm}
			\includegraphics[width=0.12\linewidth]{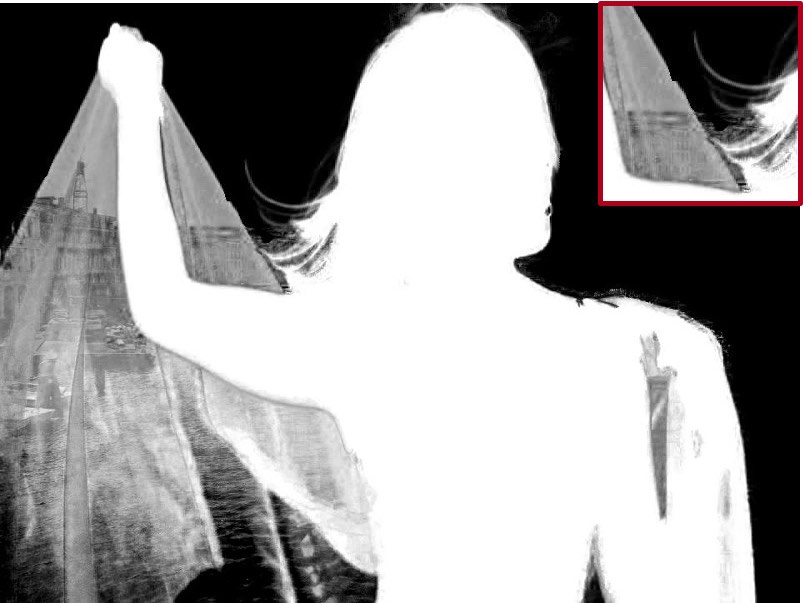} & \hspace{-0.3cm}
			\includegraphics[width=0.12\linewidth]{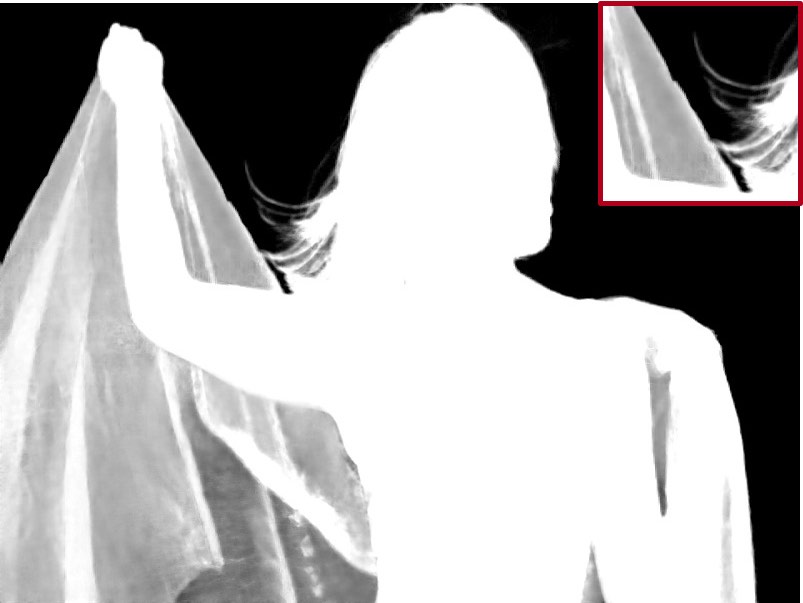} & \hspace{-0.3cm}
			\includegraphics[width=0.12\linewidth]{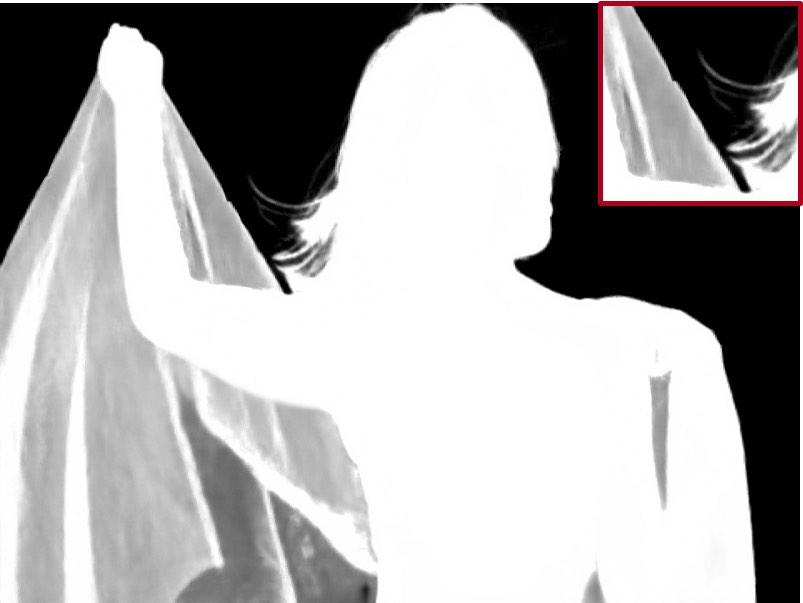} & \hspace{-0.3cm}
			\includegraphics[width=0.12\linewidth]{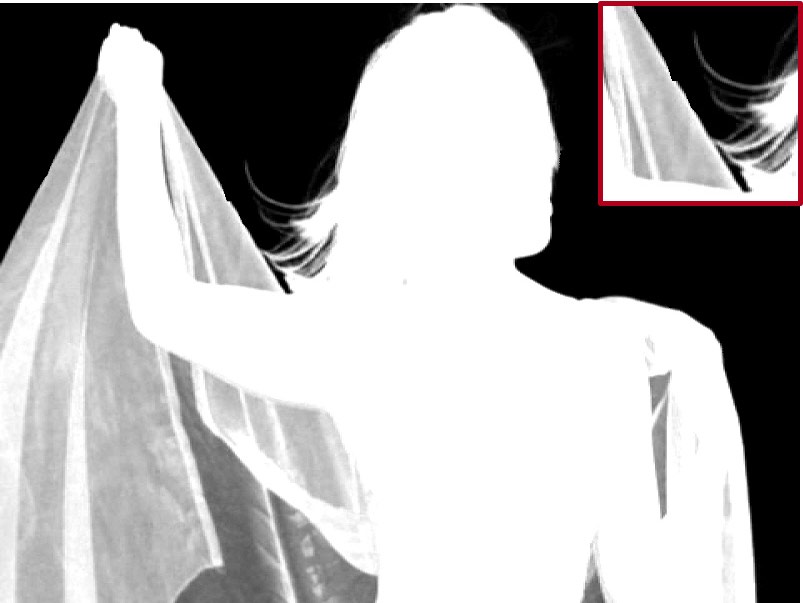}\\
			Image & PSP50 Reg & PSP50+IFM \cite{aksoy2017designing} & PSP50+DIM \cite{xu2017deep} & TrimapGT+IFM \cite{aksoy2017designing} & TrimapGT+DIM \cite{xu2017deep} & Our method & Alpha GT\\
			
			\includegraphics[width=0.12\linewidth]{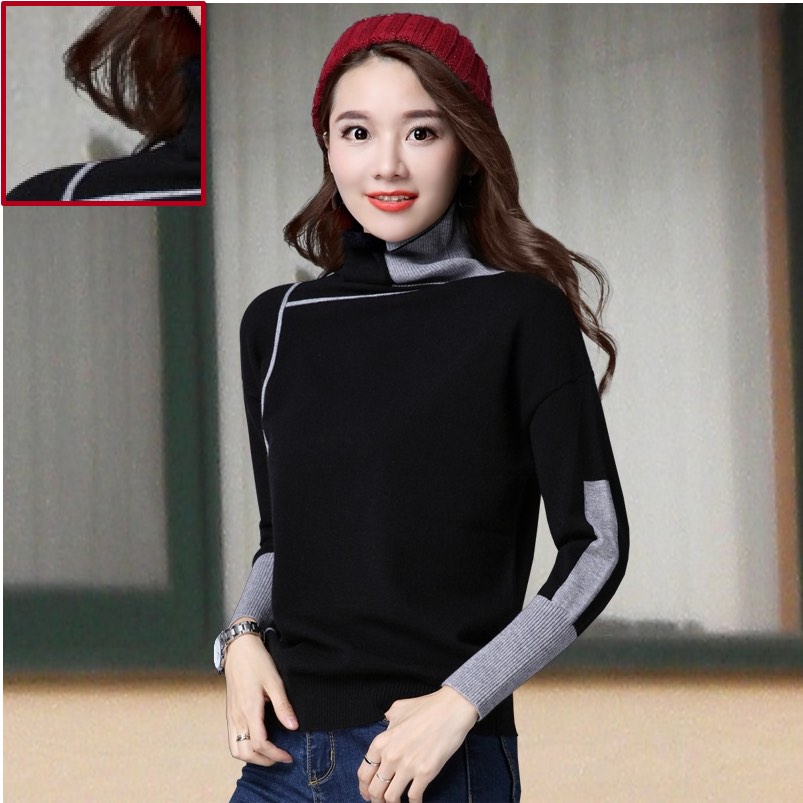} & \hspace{-0.3cm}
			\includegraphics[width=0.12\linewidth]{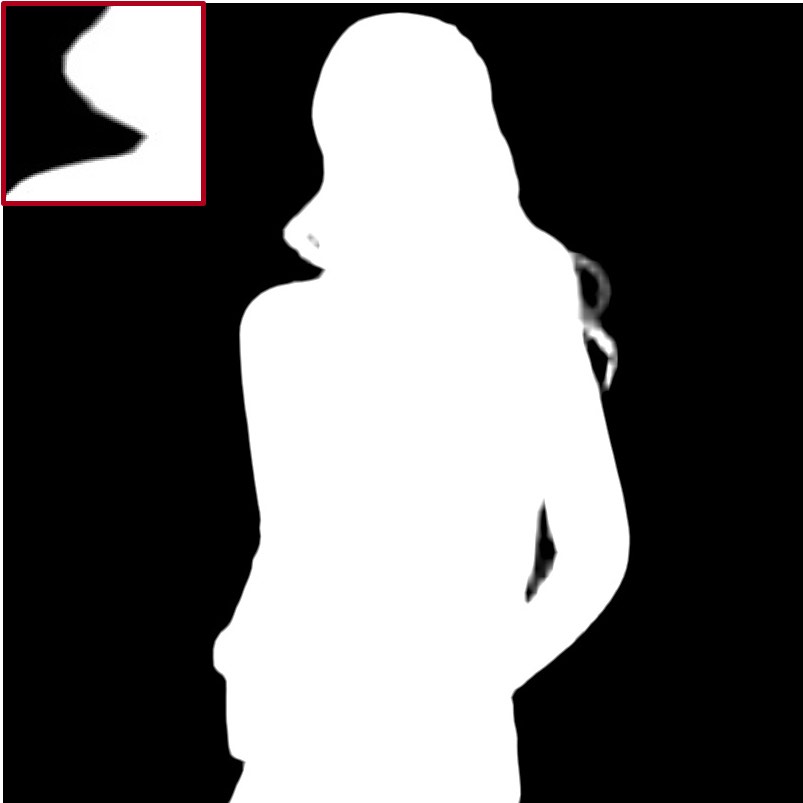} & \hspace{-0.3cm}
			\includegraphics[width=0.12\linewidth]{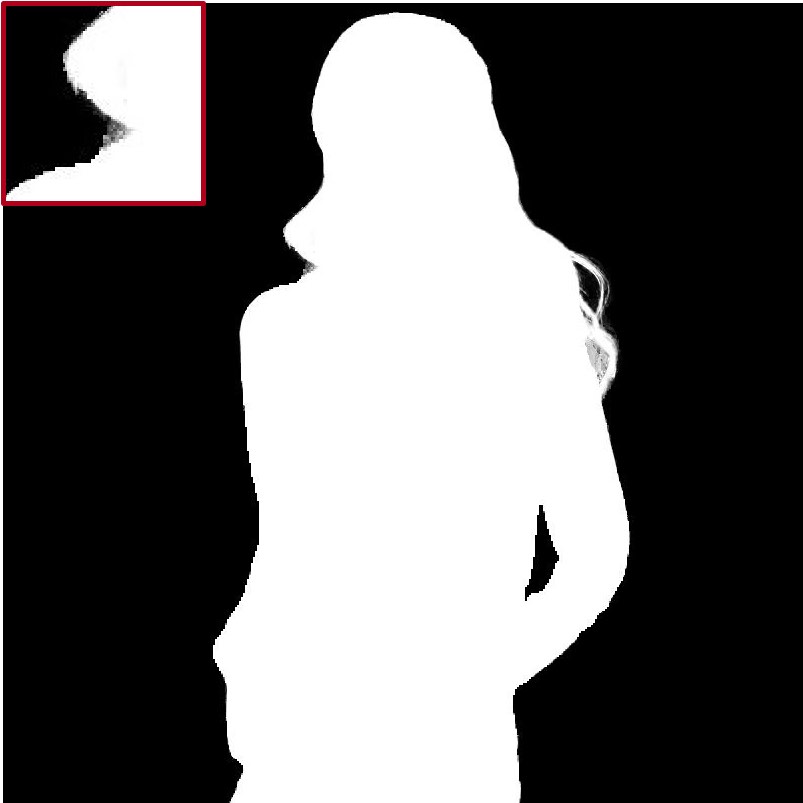} & \hspace{-0.3cm}
			\includegraphics[width=0.12\linewidth]{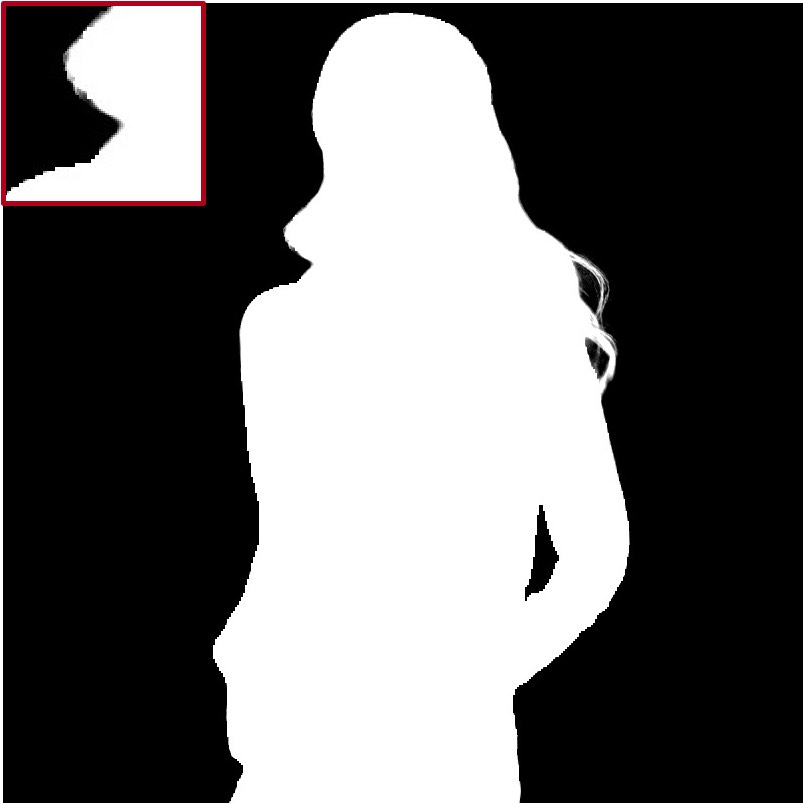} & \hspace{-0.3cm}
			\includegraphics[width=0.12\linewidth]{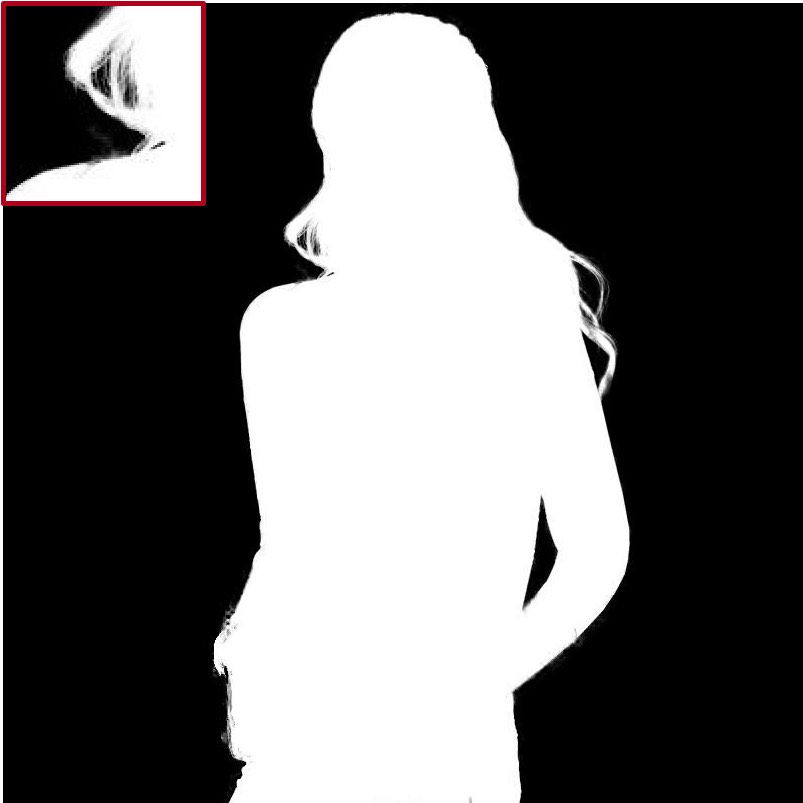} & \hspace{-0.3cm}
			\includegraphics[width=0.12\linewidth]{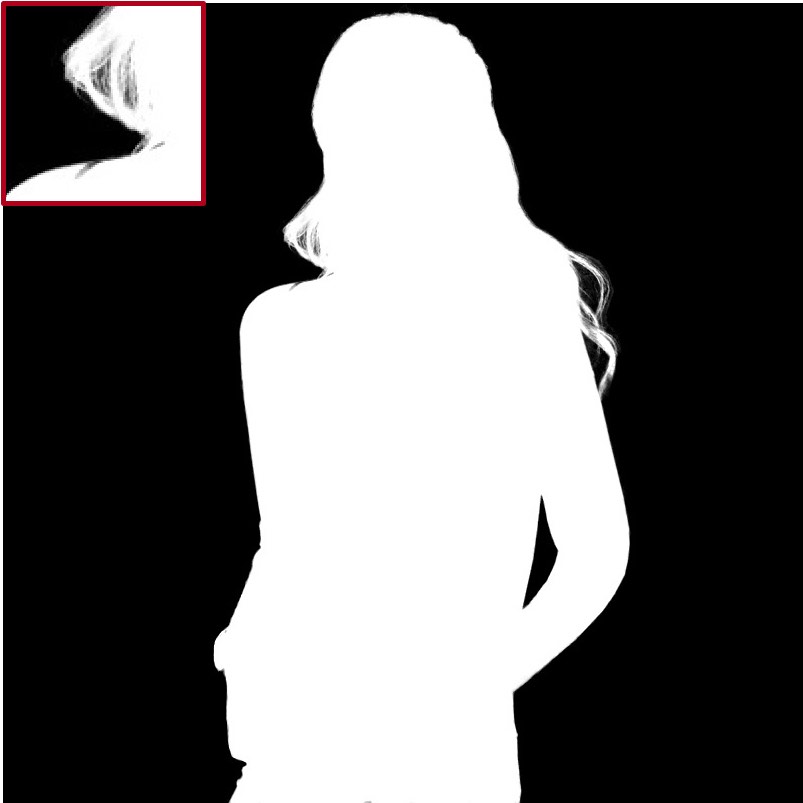} & \hspace{-0.3cm}
			\includegraphics[width=0.12\linewidth]{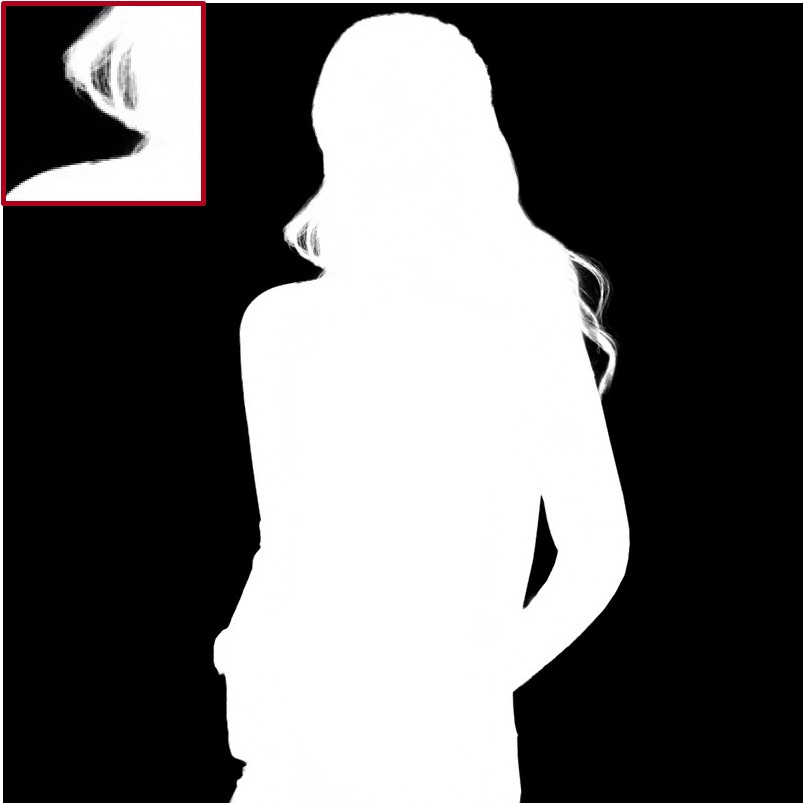} & \hspace{-0.3cm}
			\includegraphics[width=0.12\linewidth]{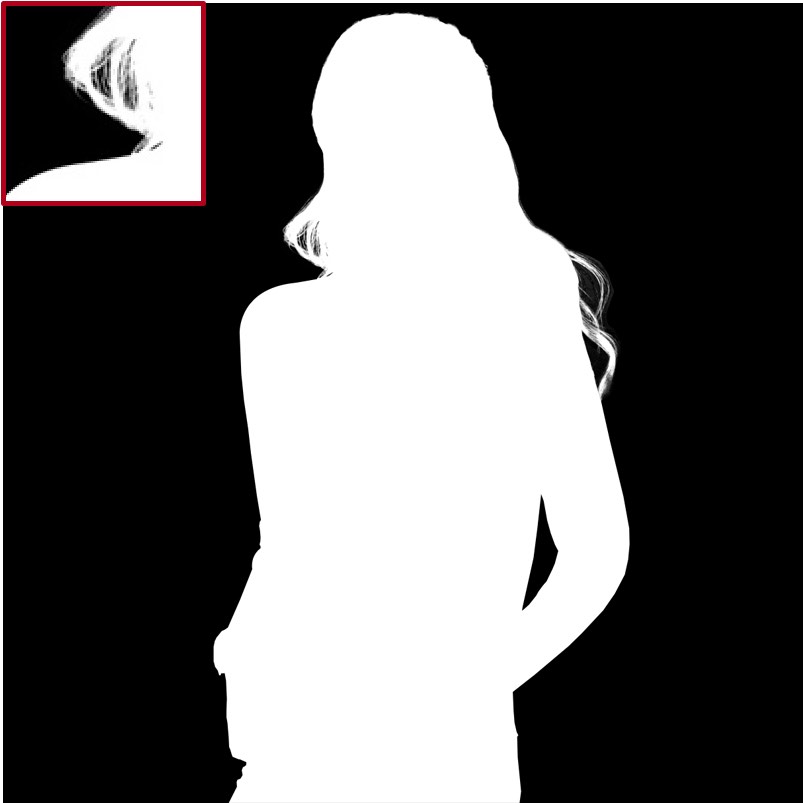}\\
			Image & PSP50 Reg & PSP50+IFM \cite{aksoy2017designing} & PSP50+DIM \cite{xu2017deep} & TrimapGT+IFM \cite{aksoy2017designing} & TrimapGT+DIM \cite{xu2017deep} & Our method & Alpha GT\\
			
		\end{tabular}
	\end{center}
	\caption{\label{fig:case_show_hmtds} The visual comparison results on the semantic human matting testing dataset}
\end{figure*}

Several visual examples are shown in Fig.~\ref{fig:case_show_hmtds}.
Compared to other methods (from column 2 to column 4), our method can not only obtain more "sharp" details, such as hairs, but also have much little semantic errors which may benefit from the end-to-end training.

%

\subsection{Automatic Method vs. Interactive Methods}

\begin{table}
	\caption{The quantitative results of our method and several state-of-the-art matting methods that need trimap on the semantic human matting testing dataset.}
	\label{tab:quan_res_hmtds_with_trimap}
	\begin{tabular}{lllll}
		\toprule
		Methods & \tabincell{c}{SAD \\ ($\times10^{-3}$)} & \tabincell{c}{MSE \\ ($\times10^{-3}$)} & \tabincell{c}{Gradient \\ ($\times10^{-5}$)} & \tabincell{c}{Connectivity \\ ($\times 10^{-5}$)} \\
		\midrule
		TrimapGT+CF & 6.772 & 2.258 & 9.0390 & 34.248\\
		TrimapGT+KNN & 8.379 & 3.413 & 16.451 & 83.458\\
		TrimapGT+DCNN & 6.760 & 2.162 & 9.753 & 44.392\\
		TrimapGT+IFM & 5.933 & 1.798 & 8.290 & 54.257\\
		TrimapGT+DIM & \textbf{2.642} & \textbf{0.589} & \textbf{3.035} & \textbf{25.773}\\
		\midrule
		Our Method & 3.833 & 1.534 & 5.179 & 36.513\\
		\bottomrule
	\end{tabular}
\end{table}

We compare our method with state-of-the-art interactive matting methods taking the groundtruth trimaps as inputs, which are generated by the same strategy used in \emph{T-Net} pretraining stage.
We denote the baselines as \emph{TrimapGT + $X$}, where $X$ represents 5 state-of-the-art matting methods including CF \cite{levin2008closed} , KNN \cite{chen2013knn}, DCNN \cite{cho2016natural}, IFM \cite{aksoy2017designing} and DIM \cite{xu2017deep}.
Table~\ref{tab:quan_res_hmtds_with_trimap} shows the comparisons.
We can see that the result of our automatic method trained by end-to-end strategy is higher than most interactive matting methods, and is slightly inferior to TrimapGT+DIM.
Note that our automatic method only takes in the original RGB images, while interactive TrimapGT + $X$ baselines take additional groundtruth trimaps as inputs.
Our \emph{T-Net} could infer the human bodies and estimate coarse predictions which are then complemented with matting details by \emph{M-Net}.
Despite slightly higher test loss, our automatic method is visually comparable with DIM, the state-of-the-art interactive matting methods, as shown in Fig.~\ref{fig:case_show_hmtds} (column "TrimapGT+DIM" 
\emph{vs.} "Our method").
 


\subsection{Evaluation and Analysis of Different Components}

\begin{table}
  \caption{Evaluation of Different Components.}
  \label{tab:ablation_study}
  \begin{tabular}{lllll}
    \toprule
    Methods & \tabincell{c}{SAD \\ ($\times10^{-3}$)} & \tabincell{c}{MSE \\ ($\times10^{-3}$)} & \tabincell{c}{Gradient \\ ($\times10^{-5}$)} & \tabincell{c}{Connectivity \\ ($\times 10^{-5}$)} \\
    \midrule
    no end-to-end & 7.576 & 4.275 & 19.762 & 52.470\\
    no Fusion & 4.231 & 2.146 & 5.230 & 56.402\\
    no $\mathcal{L}_t$ & 4.536 & 2.278 & 5.424 & 52.546 \\
    \midrule
    Our Method & \textbf{3.833} & \textbf{1.534} & \textbf{5.179} & \textbf{36.513}\\
  \bottomrule
\end{tabular}
\end{table}

\paragraph{\textbf{The Effect of End-to-end Training}}
In order to evaluate the effectiveness of the end-to-end strategy,
we compare our end-to-end trained model with that using only pre-trained parameters (\emph{no end-to-end}).
The results are listed in Table \ref{tab:ablation_study}. 
We can see that network trained in end-to-end manner performs better than \emph{no end-to-end}, which shows the effectiveness of the end-to-end training.

\paragraph{\textbf{The Evaluation of \emph{Fusion Module}}}
To validate the importance of the proposed \emph{Fusion Module}, we design a simple baseline that directly outputs the result of \emph{M-Net}, \emph{i.e.} $\alpha_p = \alpha_r$.
It is trained with the same objective as Eq.~\ref{total_loss}.
We compare the performance between our method with \emph{Fusion Module} and this baseline without \emph{Fusion Module} in Table \ref{tab:ablation_study}.
We can see that our method with \emph{Fusion Module} achieves better performance than the baseline. 
Especially note that although other metrics remain relatively small, the Connectivity error of baseline gets quite large.
It can be due to a blurring of the structural details when predicting the whole alpha matte only with \emph{M-Net}.
Thus the designed fusion module, which leverages both the coarse estimations from \emph{T-Net} and the fine predictions from M-Net, is crucial for better performance.

\paragraph{\textbf{The Effect of Constraint $\mathcal{L}_t$}}
In our implementation, we introduce a constraint for the trimap, \emph{i.e.} $\mathcal{L}_t$.
We train a network removing it to investigate the effect of such a constraint.
We denote the network trained in this way as \emph{no $\mathcal{L}_t$}.
The performance of this network is shown in Table \ref{tab:ablation_study}.
We can see that the network without $\mathcal{L}_t$ performs better than that without end-to-end training, but is worse than the proposed method.
This constraint makes the trimap more meaningful and the decomposition in $\mathcal{L}_p$ more stable.

\begin{figure*}[h!]
	\begin{center}
		\begin{tabular}{cccc}
			\includegraphics[width=0.24\linewidth]{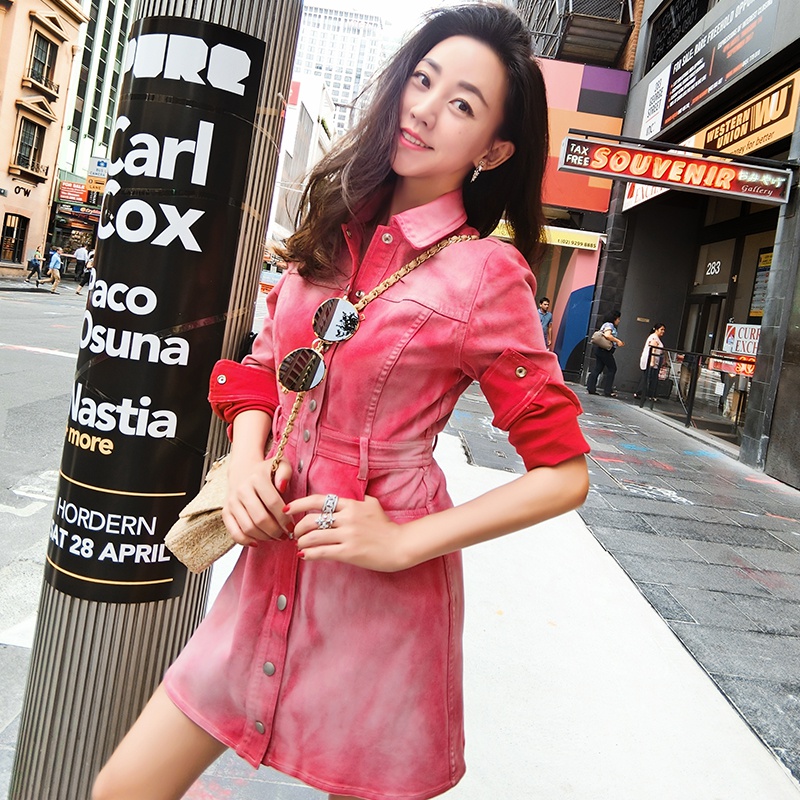} & \hspace{-0.3cm}
			\includegraphics[width=0.24\linewidth]{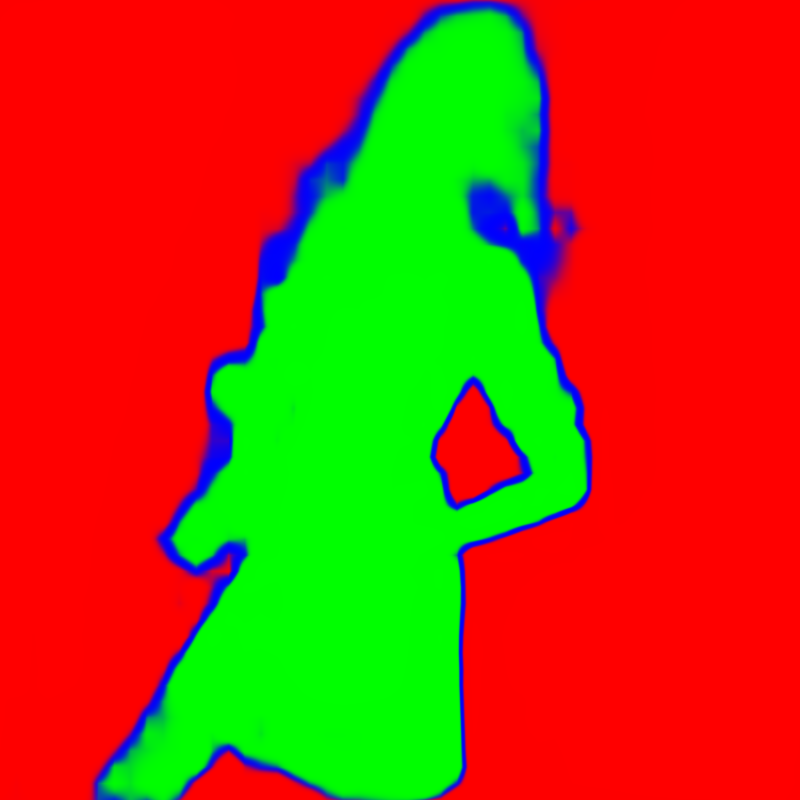} & \hspace{-0.3cm}
			\includegraphics[width=0.24\linewidth]{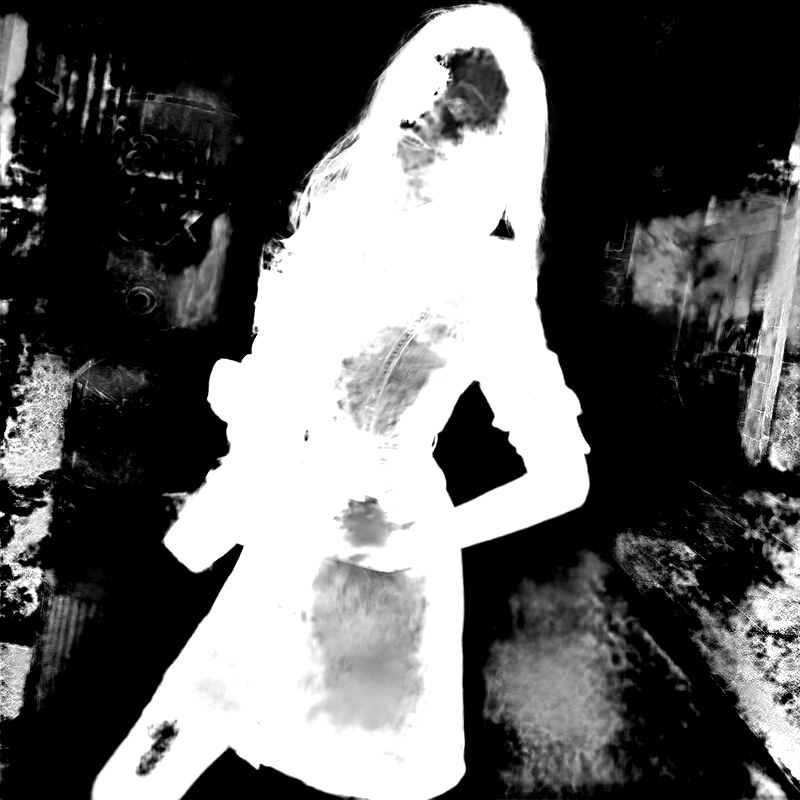} & \hspace{-0.3cm}
			\includegraphics[width=0.24\linewidth]{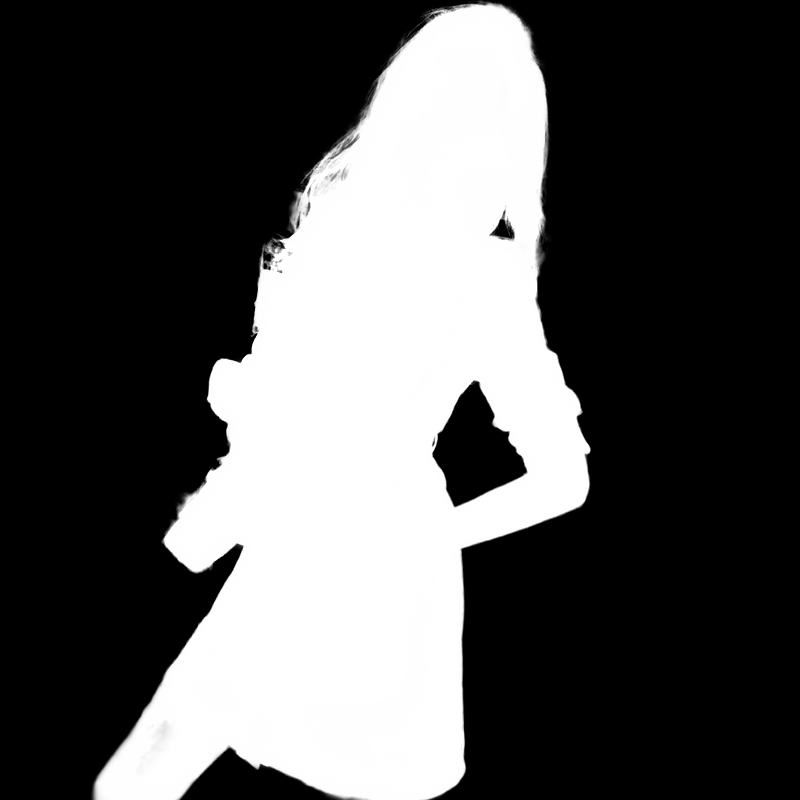}\\
			(a) & (b) & (c) & (d)\\

		\end{tabular}
	\end{center}
	\caption{\label{fig:vis} Intermediate results visualization on a real image. (a) an input image, (b) trimap predicted by T-Net, (c) raw alpha matte predicted by M-Net, (d) fusion result according to Eq. \ref{equa_t_m_fusion}.}
\end{figure*}

\begin{figure*}[h!]
\begin{center}
\begin{tabular}{cccccc}
  \includegraphics[width=0.16\linewidth]{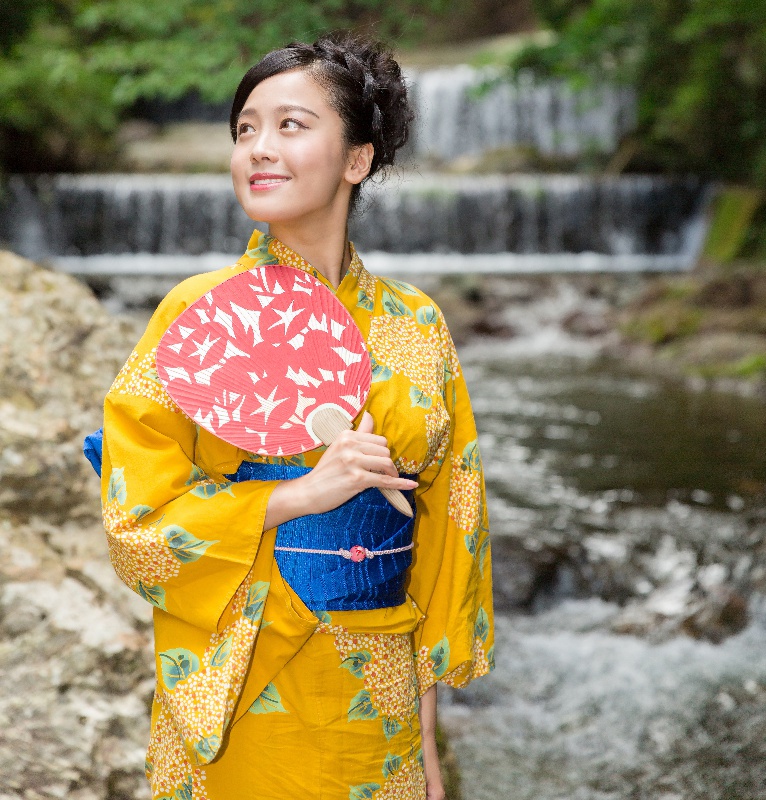} & \hspace{-0.3cm}
  \includegraphics[width=0.16\linewidth]{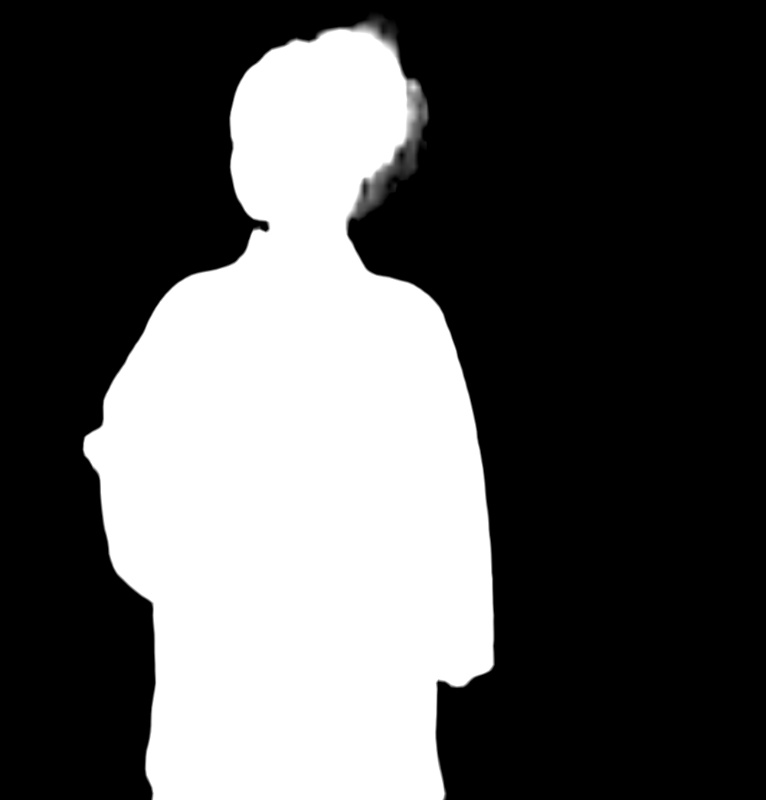} & \hspace{-0.3cm}
  \includegraphics[width=0.16\linewidth]{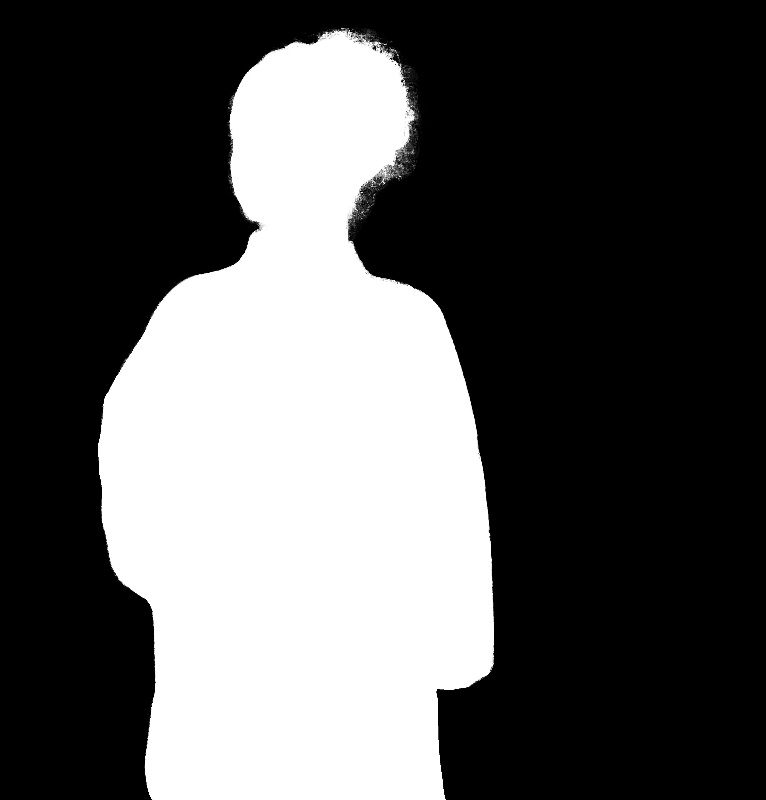} & \hspace{-0.3cm}
  \includegraphics[width=0.16\linewidth]{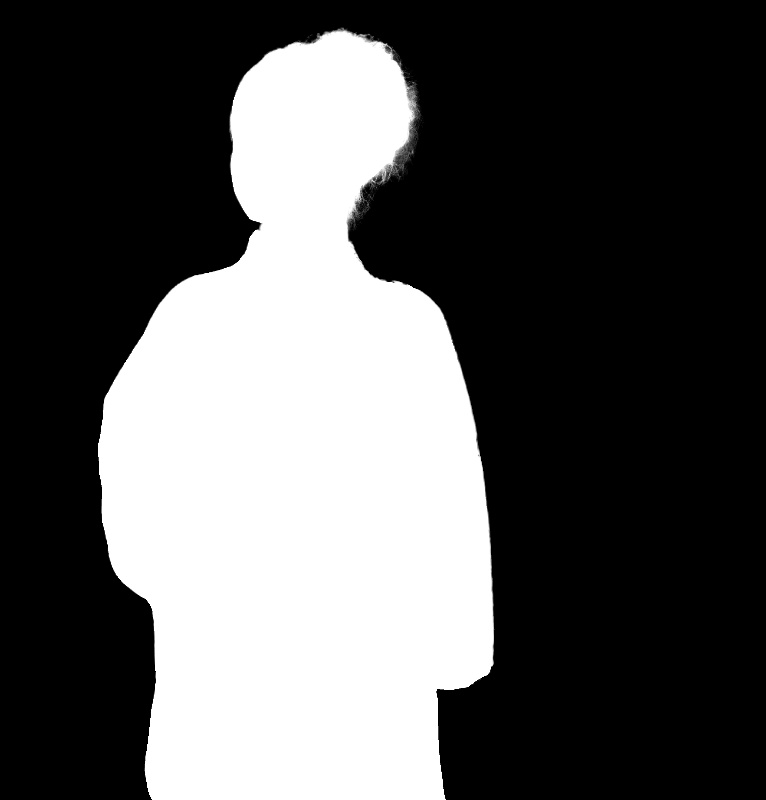} & \hspace{-0.3cm}
  \includegraphics[width=0.16\linewidth]{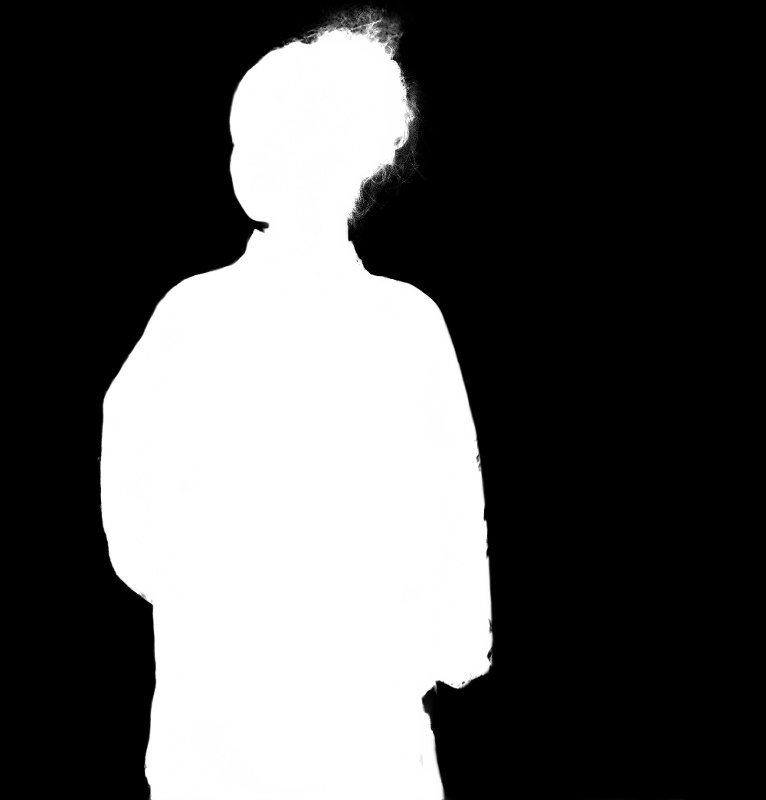} & \hspace{-0.3cm}
  \includegraphics[width=0.16\linewidth]{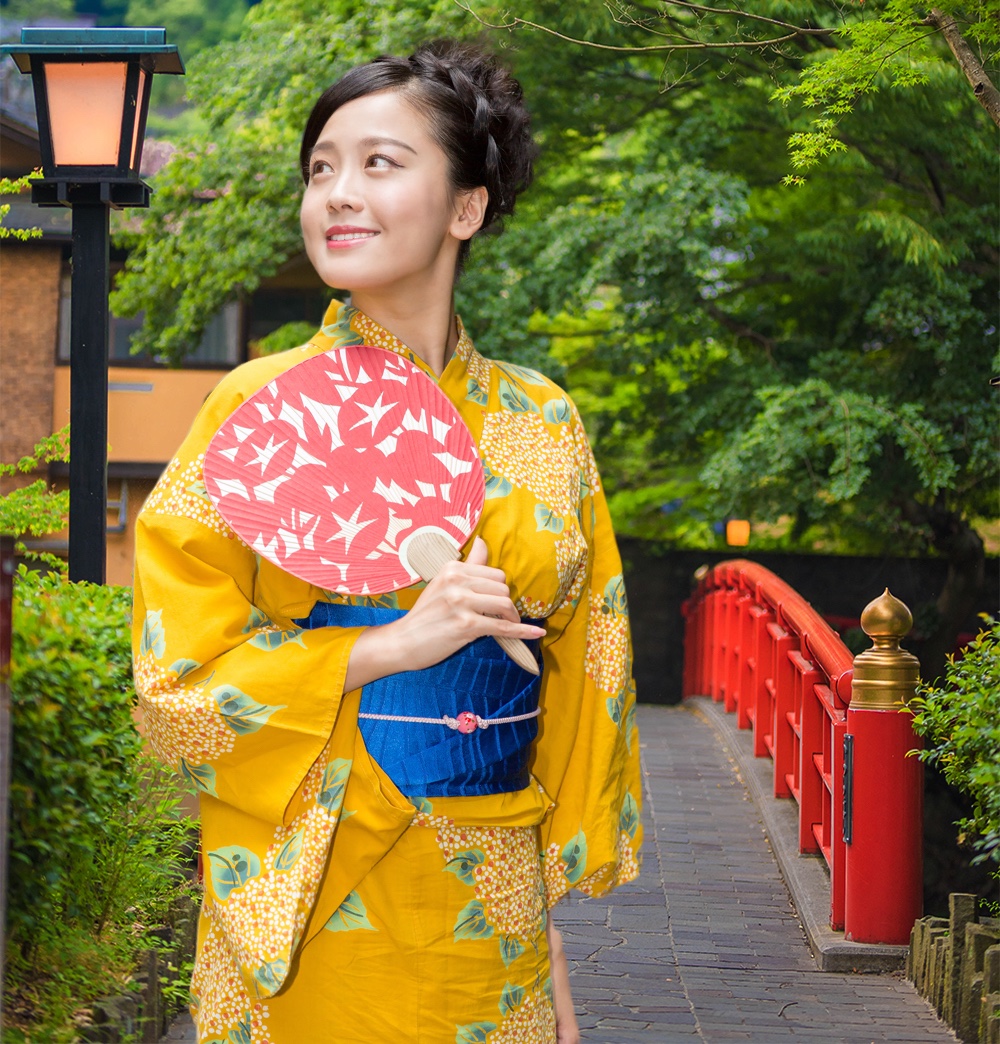}\\
  Image & PSP50 Reg & PSP50+IFM \cite{aksoy2017designing} & PSP50+DIM \cite{xu2017deep} & Our method & Composition \\
  \includegraphics[width=0.16\linewidth]{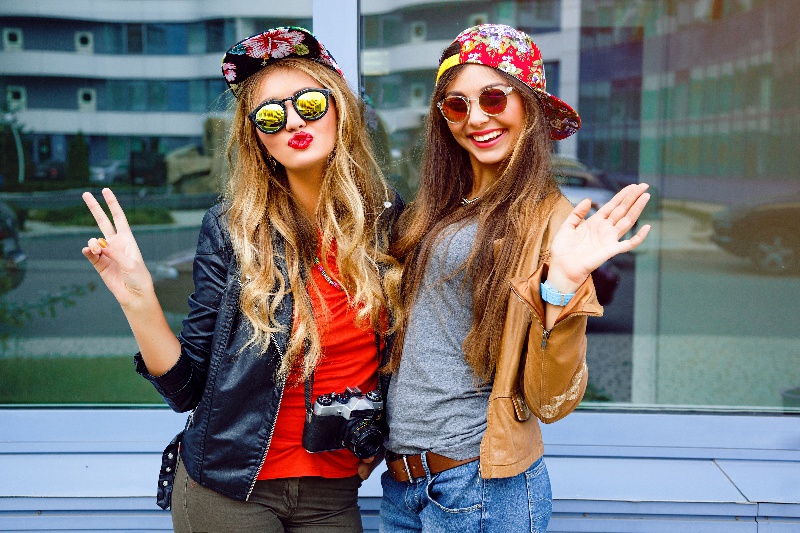} & \hspace{-0.3cm}
  \includegraphics[width=0.16\linewidth]{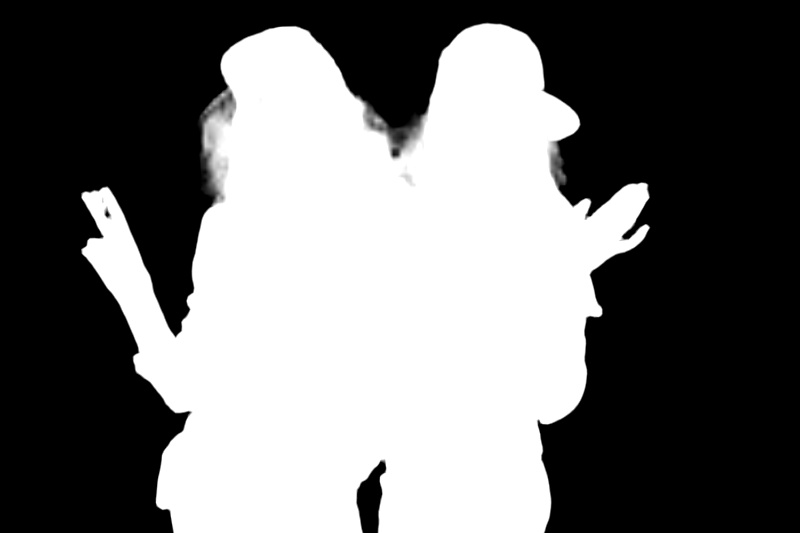} & \hspace{-0.3cm}
  \includegraphics[width=0.16\linewidth]{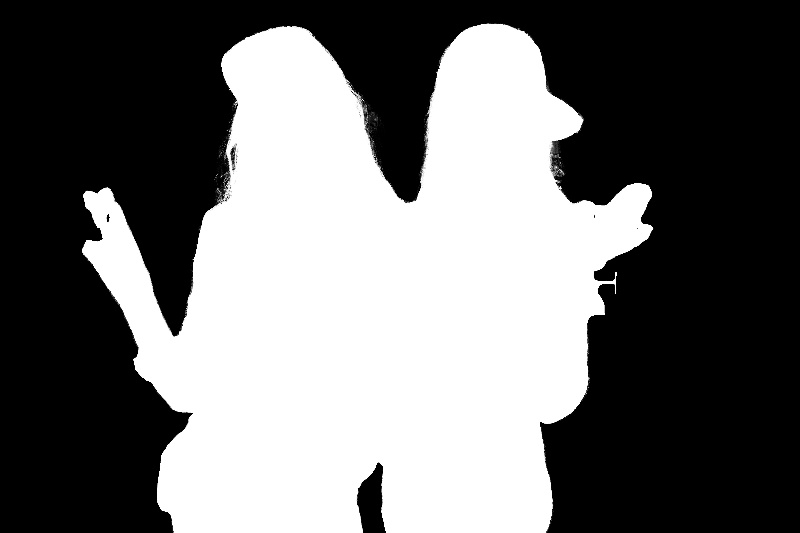} & \hspace{-0.3cm}
  \includegraphics[width=0.16\linewidth]{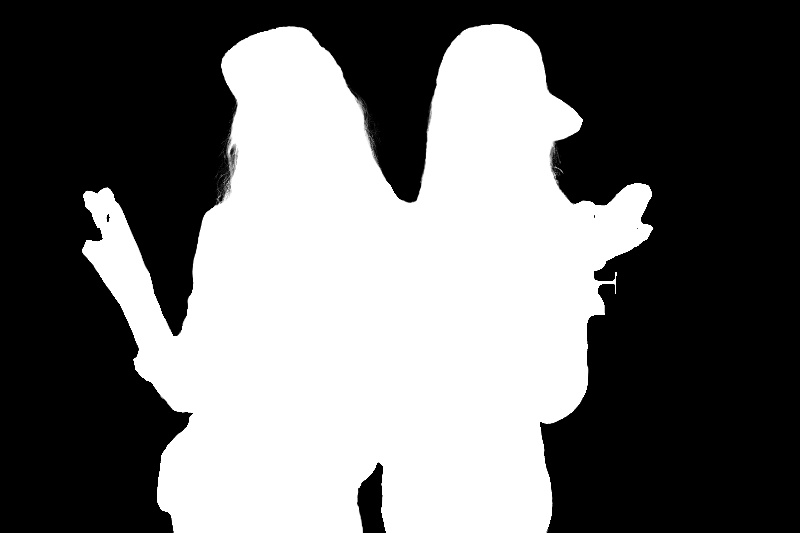} & \hspace{-0.3cm}
  \includegraphics[width=0.16\linewidth]{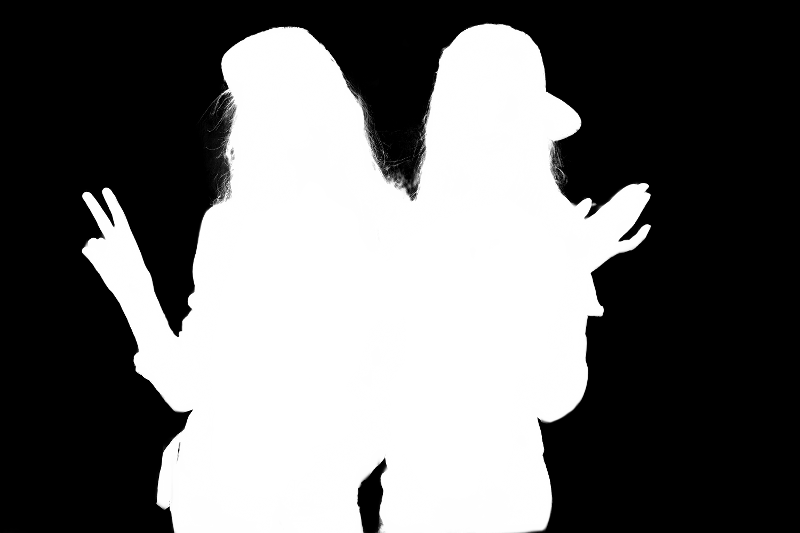} & \hspace{-0.3cm}
  \includegraphics[width=0.16\linewidth]{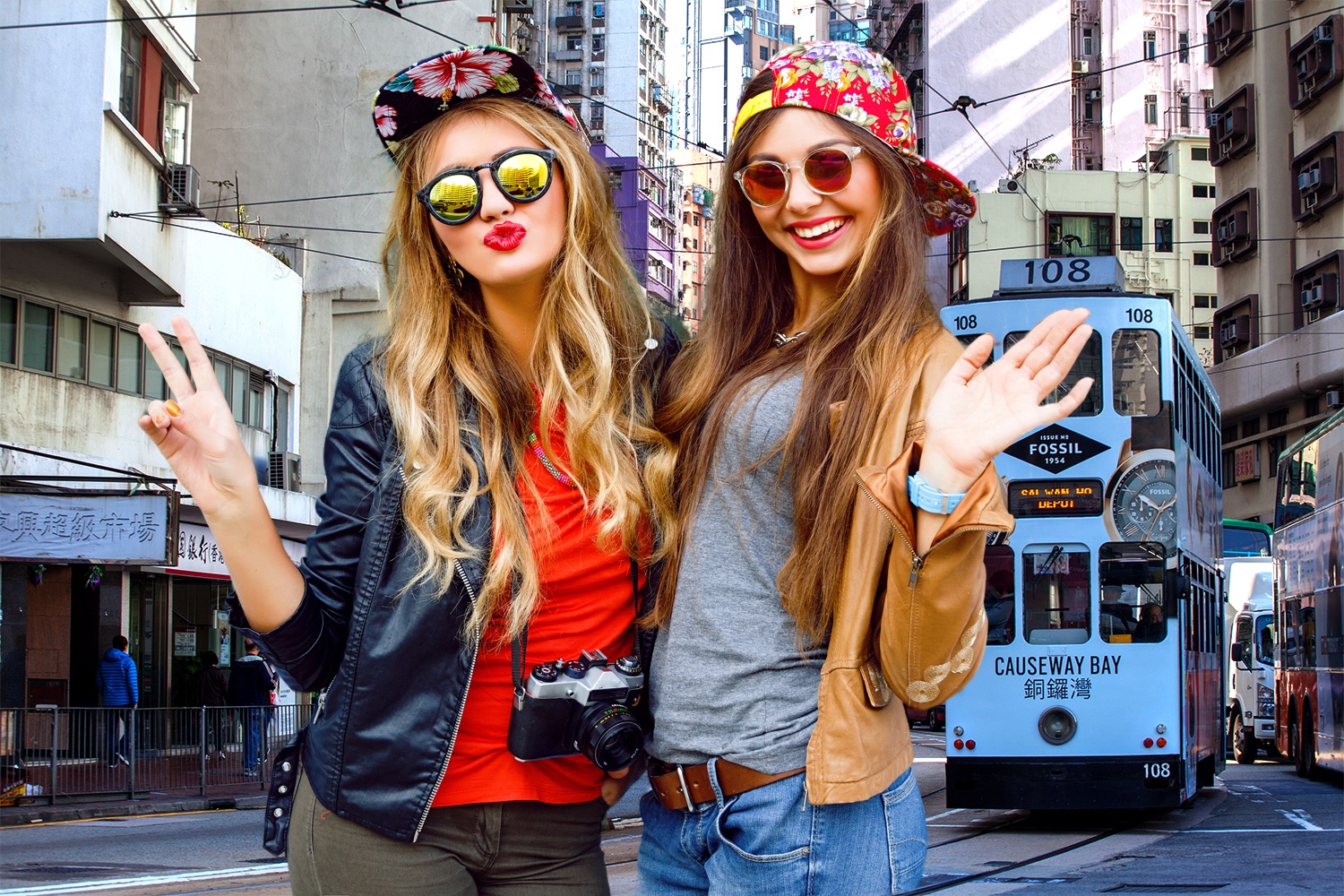}\\
  Image & PSP50 Reg & PSP50+IFM \cite{aksoy2017designing} & PSP50+DIM \cite{xu2017deep} & Our method & Composition \\   
  \includegraphics[width=0.16\linewidth]{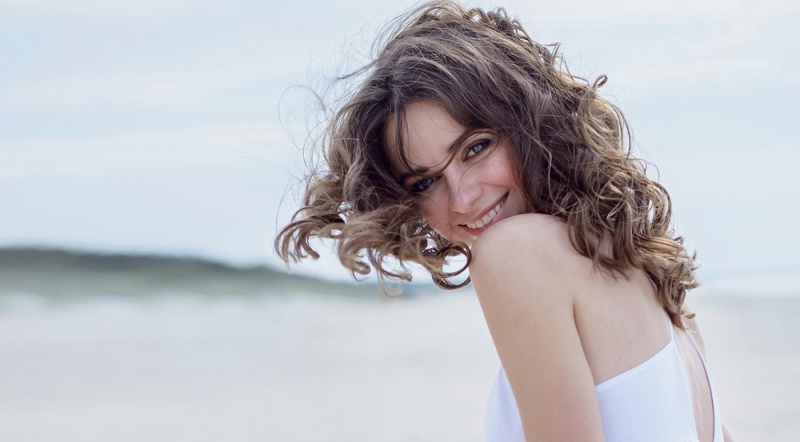} & \hspace{-0.3cm}
  \includegraphics[width=0.16\linewidth]{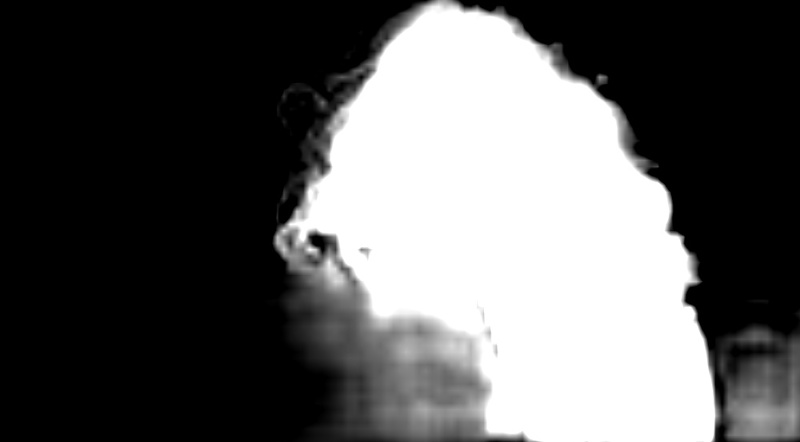} & \hspace{-0.3cm}
  \includegraphics[width=0.16\linewidth]{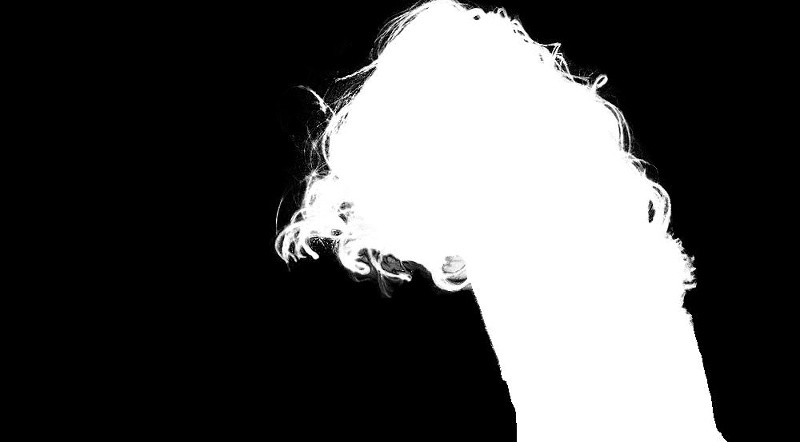} & \hspace{-0.3cm}
  \includegraphics[width=0.16\linewidth]{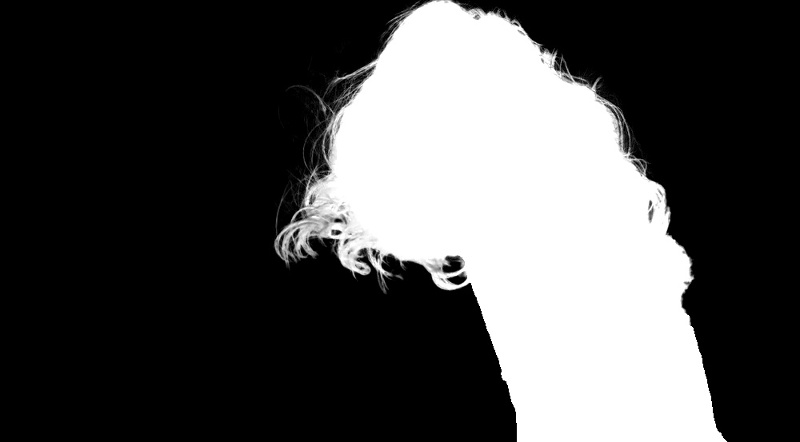} & \hspace{-0.3cm}
  \includegraphics[width=0.16\linewidth]{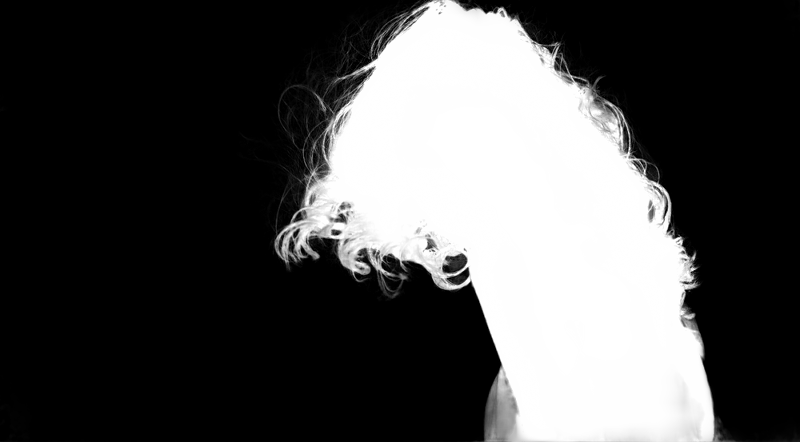} & \hspace{-0.3cm}
  \includegraphics[width=0.16\linewidth]{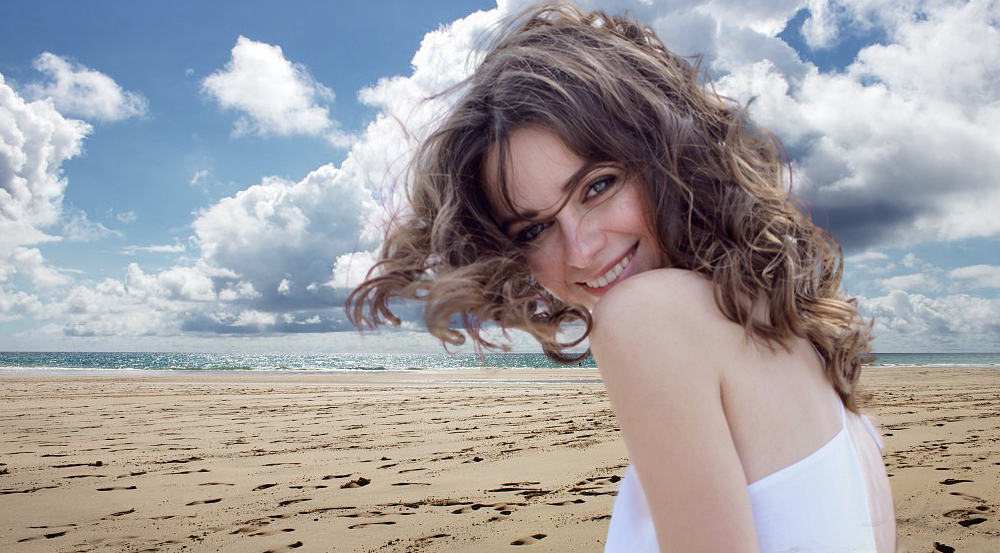}\\
  Image & PSP50 Reg & PSP50+IFM \cite{aksoy2017designing} & PSP50+DIM \cite{xu2017deep} & Our method & Composition \\   
  \includegraphics[width=0.16\linewidth]{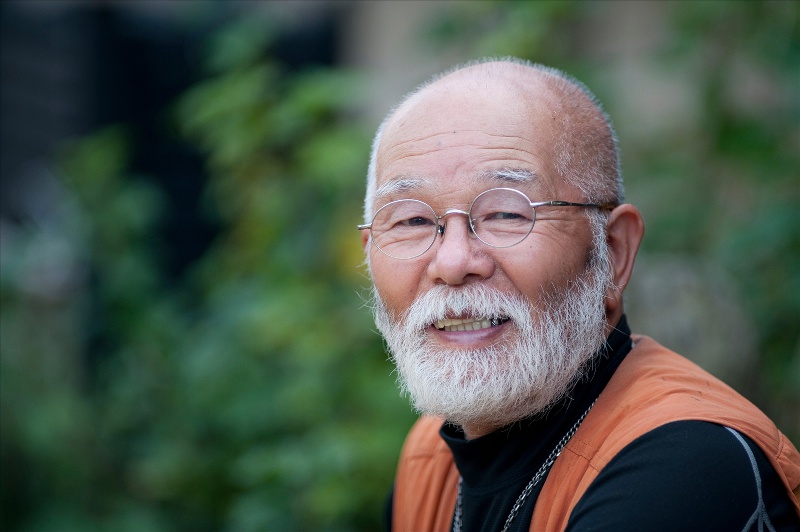} & \hspace{-0.3cm}
  \includegraphics[width=0.16\linewidth]{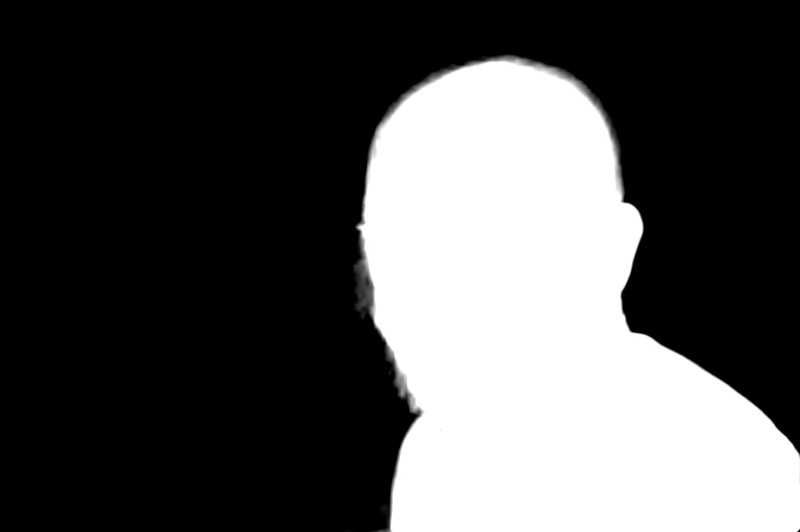} & \hspace{-0.3cm}
  \includegraphics[width=0.16\linewidth]{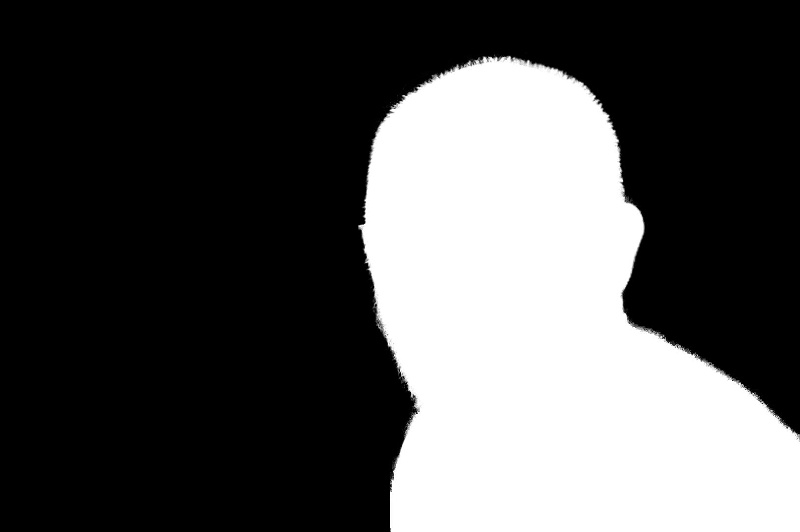} & \hspace{-0.3cm}
  \includegraphics[width=0.16\linewidth]{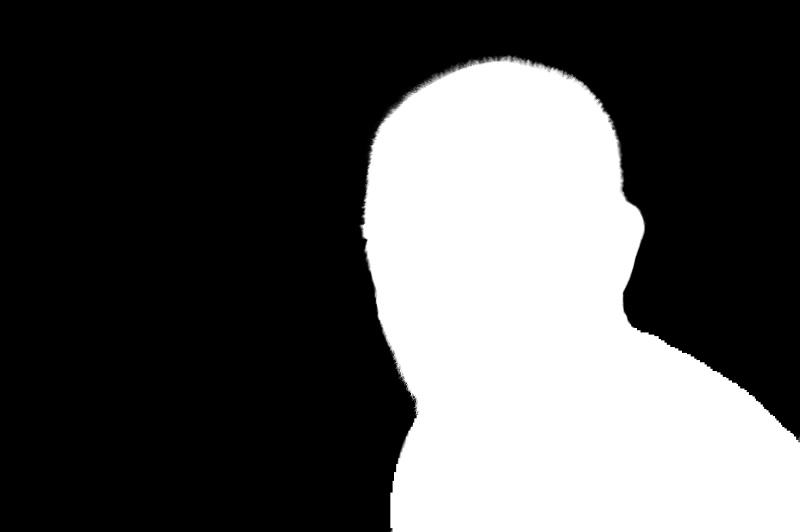} & \hspace{-0.3cm}
  \includegraphics[width=0.16\linewidth]{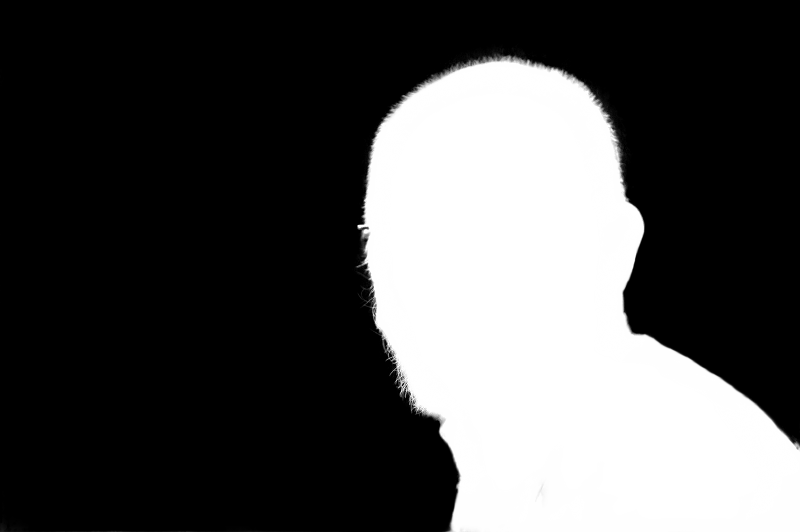} & \hspace{-0.3cm}
  \includegraphics[width=0.16\linewidth]{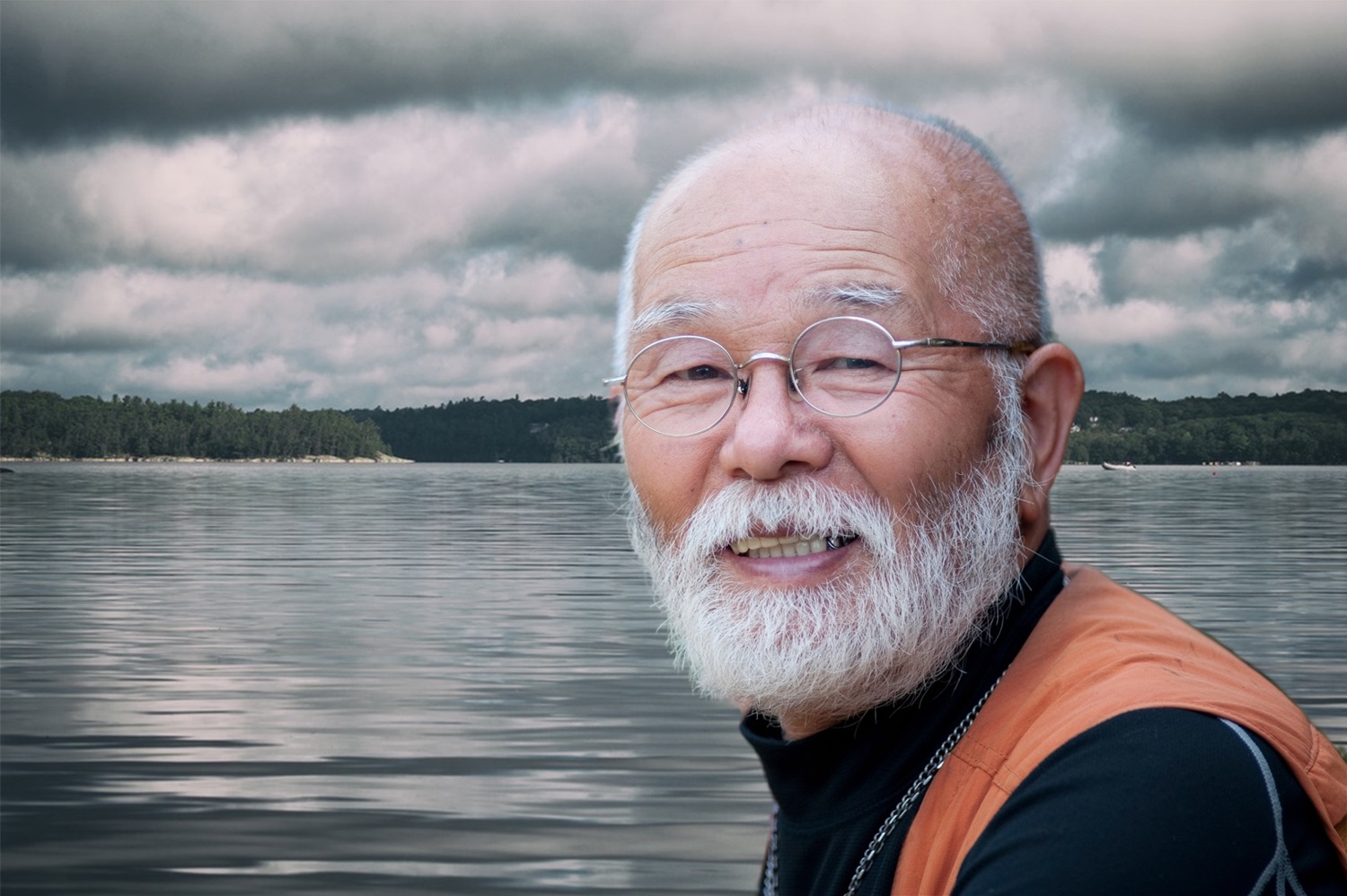} \\
  Image & PSP50 Reg & PSP50+IFM \cite{aksoy2017designing} & PSP50+DIM \cite{xu2017deep} & Our method & Composition \\   
\end{tabular}
\end{center}
\caption{\label{fig:case_show_real_images} The visual comparison results on the real images.}
\end{figure*}

\paragraph{\textbf{Visualization of Intermediate Results}}

To better understand the mechanism of SHM, we visualize the intermediate results on a real image shown in Fig \ref{fig:vis}.
The first column (a) shows the original input image, the second column (b) shows the predicted foreground (green), background (red) and unknown region (blue) from T-Net, the third column (c) shows the predicted alpha matte from M-Net, and the last column (d) shows the fusion result of the second column (b) and the third column (c) according to Eq.~\ref{equa_t_m_fusion}.
We can see that the \emph{T-Net} could segment the rough estimation of human main body, and automatically distinguish the definite human edges where predicted unknown region is narrower and structural details where predicted unknown region is wider.
In addition, with the help of the coarse prediction provided by \emph{T-Net}, \emph{M-Net} could concentrate on the transitional regions between foreground and background and predict more structural details of alpha matte.
Further, we combine the advantages of \emph{T-Net} and \emph{M-Net} and obtain a high quality alpha matte with the aid of \emph{Fusion Module}.

\subsection{Applying to real images}

Since the images in our dataset are composited with annotated foregrounds and random backgrounds, to investigate the ability of our model to generalize to real-world images,
we apply our model and other methods to plenty of real images for a qualitative analysis.
Several visual results are shown in Fig.~\ref{fig:case_show_real_images}.
We find that our method performs well on real images even with complicated backgrounds.
Note that the hair details of the woman in the first image of Fig.~\ref{fig:case_show_real_images} are only recovered nicely by our method.
Also, the fingers in the second image are blurred incorrectly by other methods, whereas our method distinguishes them well.
Compsition examples of the foregrounds and new backgrounds with the help of automatically predicted alpha matte are illustrated in the last column of Fig.~\ref{fig:case_show_real_images}.
We can see these compositions have high visual quality.
More results can be found in supplementary materials.

\section{Conclusion}

In this paper, we focus on the human matting problem which shows a great importance for a wide variety of applications. 
In order to simultaneously capture global semantic information and local details, we propose to cascade a trimap network and a matting network, as well as a novel fusion module to generate alpha matte automatically.
Furthermore, we create a large high quality human matting dataset.
Benefitting from the model structure and dataset, 
our automatic human matting achieves comparable results with state-of-the-art interactive matting methods.

\section*{Acknowledgement}

We thank Jian Xu for many helpful discussions and valuable suggestions, and Yangjian Chen, Xiaowei Li, Hui Chen, Yuqi Chen for their support on developing the image labeling tool, and Min Zhou for some comments that improved the manuscript.

%% file: ms.bbl

%% file: ms.bbl
\begin{thebibliography}{30}


\ifx \showCODEN    \undefined \def \showCODEN     #1{\unskip}     \fi
\ifx \showDOI      \undefined \def \showDOI       #1{#1}\fi
\ifx \showISBNx    \undefined \def \showISBNx     #1{\unskip}     \fi
\ifx \showISBNxiii \undefined \def \showISBNxiii  #1{\unskip}     \fi
\ifx \showISSN     \undefined \def \showISSN      #1{\unskip}     \fi
\ifx \showLCCN     \undefined \def \showLCCN      #1{\unskip}     \fi
\ifx \shownote     \undefined \def \shownote      #1{#1}          \fi
\ifx \showarticletitle \undefined \def \showarticletitle #1{#1}   \fi
\ifx \showURL      \undefined \def \showURL       {\relax}        \fi
\providecommand\bibfield[2]{#2}
\providecommand\bibinfo[2]{#2}
\providecommand\natexlab[1]{#1}
\providecommand\showeprint[2][]{arXiv:#2}

\bibitem[\protect\citeauthoryear{Aksoy, Ayd{\i}n, Pollefeys, and
  Z{\"u}rich}{Aksoy et~al\mbox{.}}{2017}]%
        {aksoy2017designing}
\bibfield{author}{\bibinfo{person}{Yag{\i}z Aksoy},
  \bibinfo{person}{Tun{\c{c}}~Ozan Ayd{\i}n}, \bibinfo{person}{Marc Pollefeys},
  {and} \bibinfo{person}{ETH Z{\"u}rich}.} \bibinfo{year}{2017}\natexlab{}.
\newblock \showarticletitle{Designing effective inter-pixel information flow
  for natural image matting}. In \bibinfo{booktitle}{\emph{Computer Vision and
  Pattern Recognition (CVPR)}}.
\newblock


\bibitem[\protect\citeauthoryear{Butler, Wulff, Stanley, and Black}{Butler
  et~al\mbox{.}}{2012}]%
        {butler2012naturalistic}
\bibfield{author}{\bibinfo{person}{Daniel~J Butler}, \bibinfo{person}{Jonas
  Wulff}, \bibinfo{person}{Garrett~B Stanley}, {and} \bibinfo{person}{Michael~J
  Black}.} \bibinfo{year}{2012}\natexlab{}.
\newblock \showarticletitle{A naturalistic open source movie for optical flow
  evaluation}. In \bibinfo{booktitle}{\emph{European Conference on Computer
  Vision}}. Springer, \bibinfo{pages}{611--625}.
\newblock


\bibitem[\protect\citeauthoryear{Chen, Papandreou, Kokkinos, Murphy, and
  Yuille}{Chen et~al\mbox{.}}{2016}]%
        {chen2016deeplab}
\bibfield{author}{\bibinfo{person}{Liang-Chieh Chen}, \bibinfo{person}{George
  Papandreou}, \bibinfo{person}{Iasonas Kokkinos}, \bibinfo{person}{Kevin
  Murphy}, {and} \bibinfo{person}{Alan~L Yuille}.}
  \bibinfo{year}{2016}\natexlab{}.
\newblock \showarticletitle{Deeplab: Semantic image segmentation with deep
  convolutional nets, atrous convolution, and fully connected crfs}.
\newblock \bibinfo{journal}{\emph{arXiv preprint arXiv:1606.00915}}
  (\bibinfo{year}{2016}).
\newblock


\bibitem[\protect\citeauthoryear{Chen, Li, and Tang}{Chen
  et~al\mbox{.}}{2013}]%
        {chen2013knn}
\bibfield{author}{\bibinfo{person}{Qifeng Chen}, \bibinfo{person}{Dingzeyu Li},
  {and} \bibinfo{person}{Chi-Keung Tang}.} \bibinfo{year}{2013}\natexlab{}.
\newblock \showarticletitle{KNN matting}.
\newblock \bibinfo{journal}{\emph{IEEE transactions on pattern analysis and
  machine intelligence}} \bibinfo{volume}{35}, \bibinfo{number}{9}
  (\bibinfo{year}{2013}), \bibinfo{pages}{2175--2188}.
\newblock


\bibitem[\protect\citeauthoryear{Cho, Tai, and Kweon}{Cho
  et~al\mbox{.}}{2016}]%
        {cho2016natural}
\bibfield{author}{\bibinfo{person}{Donghyeon Cho}, \bibinfo{person}{Yu-Wing
  Tai}, {and} \bibinfo{person}{Inso Kweon}.} \bibinfo{year}{2016}\natexlab{}.
\newblock \showarticletitle{Natural image matting using deep convolutional
  neural networks}. In \bibinfo{booktitle}{\emph{European Conference on
  Computer Vision}}. Springer, \bibinfo{pages}{626--643}.
\newblock


\bibitem[\protect\citeauthoryear{Chuang, Curless, Salesin, and Szeliski}{Chuang
  et~al\mbox{.}}{2001}]%
        {chuang2001bayesian}
\bibfield{author}{\bibinfo{person}{Yung-Yu Chuang}, \bibinfo{person}{Brian
  Curless}, \bibinfo{person}{David~H Salesin}, {and} \bibinfo{person}{Richard
  Szeliski}.} \bibinfo{year}{2001}\natexlab{}.
\newblock \showarticletitle{A bayesian approach to digital matting}. In
  \bibinfo{booktitle}{\emph{Computer Vision and Pattern Recognition, 2001. CVPR
  2001. Proceedings of the 2001 IEEE Computer Society Conference on}},
  Vol.~\bibinfo{volume}{2}. IEEE, \bibinfo{pages}{II--II}.
\newblock


\bibitem[\protect\citeauthoryear{Everingham, Van~Gool, Williams, Winn, and
  Zisserman}{Everingham et~al\mbox{.}}{[n. d.]}]%
        {pascal-voc-2012}
\bibfield{author}{\bibinfo{person}{M. Everingham}, \bibinfo{person}{L.
  Van~Gool}, \bibinfo{person}{C.~K.~I. Williams}, \bibinfo{person}{J. Winn},
  {and} \bibinfo{person}{A. Zisserman}.} \bibinfo{year}{[n. d.]}\natexlab{}.
\newblock \bibinfo{title}{The {PASCAL} {V}isual {O}bject {C}lasses {C}hallenge
  2012 {(VOC2012)} {R}esults}.
\newblock
  \bibinfo{howpublished}{http://www.pascal-network.org/challenges/VOC/voc2012/workshop/index.html}.
\newblock


\bibitem[\protect\citeauthoryear{Gastal and Oliveira}{Gastal and
  Oliveira}{2010}]%
        {gastal2010shared}
\bibfield{author}{\bibinfo{person}{Eduardo~SL Gastal} {and}
  \bibinfo{person}{Manuel~M Oliveira}.} \bibinfo{year}{2010}\natexlab{}.
\newblock \showarticletitle{Shared Sampling for Real-Time Alpha Matting}. In
  \bibinfo{booktitle}{\emph{Computer Graphics Forum}},
  Vol.~\bibinfo{volume}{29}. Wiley Online Library, \bibinfo{pages}{575--584}.
\newblock


\bibitem[\protect\citeauthoryear{Grady, Schiwietz, Aharon, and
  Westermann}{Grady et~al\mbox{.}}{2005}]%
        {grady2005random}
\bibfield{author}{\bibinfo{person}{Leo Grady}, \bibinfo{person}{Thomas
  Schiwietz}, \bibinfo{person}{Shmuel Aharon}, {and}
  \bibinfo{person}{R{\"u}diger Westermann}.} \bibinfo{year}{2005}\natexlab{}.
\newblock \showarticletitle{Random walks for interactive alpha-matting}. In
  \bibinfo{booktitle}{\emph{Proceedings of VIIP}}, Vol.~\bibinfo{volume}{2005}.
  \bibinfo{pages}{423--429}.
\newblock


\bibitem[\protect\citeauthoryear{Gupta, Vedaldi, and Zisserman}{Gupta
  et~al\mbox{.}}{2016}]%
        {gupta2016synthetic}
\bibfield{author}{\bibinfo{person}{Ankush Gupta}, \bibinfo{person}{Andrea
  Vedaldi}, {and} \bibinfo{person}{Andrew Zisserman}.}
  \bibinfo{year}{2016}\natexlab{}.
\newblock \showarticletitle{Synthetic data for text localisation in natural
  images}. In \bibinfo{booktitle}{\emph{Proceedings of the IEEE Conference on
  Computer Vision and Pattern Recognition}}. \bibinfo{pages}{2315--2324}.
\newblock


\bibitem[\protect\citeauthoryear{He, Rhemann, Rother, Tang, and Sun}{He
  et~al\mbox{.}}{2011}]%
        {he2011global}
\bibfield{author}{\bibinfo{person}{Kaiming He}, \bibinfo{person}{Christoph
  Rhemann}, \bibinfo{person}{Carsten Rother}, \bibinfo{person}{Xiaoou Tang},
  {and} \bibinfo{person}{Jian Sun}.} \bibinfo{year}{2011}\natexlab{}.
\newblock \showarticletitle{A global sampling method for alpha matting}. In
  \bibinfo{booktitle}{\emph{Computer Vision and Pattern Recognition (CVPR),
  2011 IEEE Conference on}}. IEEE, \bibinfo{pages}{2049--2056}.
\newblock


\bibitem[\protect\citeauthoryear{He, Sun, and Tang}{He et~al\mbox{.}}{2010}]%
        {he2010guided}
\bibfield{author}{\bibinfo{person}{Kaiming He}, \bibinfo{person}{Jian Sun},
  {and} \bibinfo{person}{Xiaoou Tang}.} \bibinfo{year}{2010}\natexlab{}.
\newblock \showarticletitle{Guided image filtering}. In
  \bibinfo{booktitle}{\emph{European conference on computer vision}}. Springer,
  \bibinfo{pages}{1--14}.
\newblock


\bibitem[\protect\citeauthoryear{He, Zhang, Ren, and Sun}{He
  et~al\mbox{.}}{2016}]%
        {he2016deep}
\bibfield{author}{\bibinfo{person}{Kaiming He}, \bibinfo{person}{Xiangyu
  Zhang}, \bibinfo{person}{Shaoqing Ren}, {and} \bibinfo{person}{Jian Sun}.}
  \bibinfo{year}{2016}\natexlab{}.
\newblock \showarticletitle{Deep residual learning for image recognition}. In
  \bibinfo{booktitle}{\emph{Proceedings of the IEEE conference on computer
  vision and pattern recognition}}. \bibinfo{pages}{770--778}.
\newblock


\bibitem[\protect\citeauthoryear{Hinton, Osindero, and Teh}{Hinton
  et~al\mbox{.}}{2006}]%
        {hinton2006fast}
\bibfield{author}{\bibinfo{person}{Geoffrey~E Hinton}, \bibinfo{person}{Simon
  Osindero}, {and} \bibinfo{person}{Yee-Whye Teh}.}
  \bibinfo{year}{2006}\natexlab{}.
\newblock \showarticletitle{A fast learning algorithm for deep belief nets}.
\newblock \bibinfo{journal}{\emph{Neural computation}} \bibinfo{volume}{18},
  \bibinfo{number}{7} (\bibinfo{year}{2006}), \bibinfo{pages}{1527--1554}.
\newblock


\bibitem[\protect\citeauthoryear{Lee and Wu}{Lee and Wu}{2011}]%
        {lee2011nonlocal}
\bibfield{author}{\bibinfo{person}{Philip Lee} {and} \bibinfo{person}{Ying
  Wu}.} \bibinfo{year}{2011}\natexlab{}.
\newblock \showarticletitle{Nonlocal matting}. In
  \bibinfo{booktitle}{\emph{Computer Vision and Pattern Recognition (CVPR),
  2011 IEEE Conference on}}. IEEE, \bibinfo{pages}{2193--2200}.
\newblock


\bibitem[\protect\citeauthoryear{Levin, Lischinski, and Weiss}{Levin
  et~al\mbox{.}}{2008}]%
        {levin2008closed}
\bibfield{author}{\bibinfo{person}{Anat Levin}, \bibinfo{person}{Dani
  Lischinski}, {and} \bibinfo{person}{Yair Weiss}.}
  \bibinfo{year}{2008}\natexlab{}.
\newblock \showarticletitle{A closed-form solution to natural image matting}.
\newblock \bibinfo{journal}{\emph{IEEE Transactions on Pattern Analysis and
  Machine Intelligence}} \bibinfo{volume}{30}, \bibinfo{number}{2}
  (\bibinfo{year}{2008}), \bibinfo{pages}{228--242}.
\newblock


\bibitem[\protect\citeauthoryear{Lin, Maire, Belongie, Hays, Perona, Ramanan,
  Doll{\'a}r, and Zitnick}{Lin et~al\mbox{.}}{2014}]%
        {lin2014microsoft}
\bibfield{author}{\bibinfo{person}{Tsung-Yi Lin}, \bibinfo{person}{Michael
  Maire}, \bibinfo{person}{Serge Belongie}, \bibinfo{person}{James Hays},
  \bibinfo{person}{Pietro Perona}, \bibinfo{person}{Deva Ramanan},
  \bibinfo{person}{Piotr Doll{\'a}r}, {and} \bibinfo{person}{C~Lawrence
  Zitnick}.} \bibinfo{year}{2014}\natexlab{}.
\newblock \showarticletitle{Microsoft coco: Common objects in context}. In
  \bibinfo{booktitle}{\emph{European conference on computer vision}}. Springer,
  \bibinfo{pages}{740--755}.
\newblock


\bibitem[\protect\citeauthoryear{Long, Shelhamer, and Darrell}{Long
  et~al\mbox{.}}{2015}]%
        {long2015fully}
\bibfield{author}{\bibinfo{person}{Jonathan Long}, \bibinfo{person}{Evan
  Shelhamer}, {and} \bibinfo{person}{Trevor Darrell}.}
  \bibinfo{year}{2015}\natexlab{}.
\newblock \showarticletitle{Fully convolutional networks for semantic
  segmentation}. In \bibinfo{booktitle}{\emph{Proceedings of the IEEE
  conference on computer vision and pattern recognition}}.
  \bibinfo{pages}{3431--3440}.
\newblock


\bibitem[\protect\citeauthoryear{Paszke, Gross, Chintala, Chanan, Yang, DeVito,
  Lin, Desmaison, Antiga, and Lerer}{Paszke et~al\mbox{.}}{2017}]%
        {paszke2017automatic}
\bibfield{author}{\bibinfo{person}{Adam Paszke}, \bibinfo{person}{Sam Gross},
  \bibinfo{person}{Soumith Chintala}, \bibinfo{person}{Gregory Chanan},
  \bibinfo{person}{Edward Yang}, \bibinfo{person}{Zachary DeVito},
  \bibinfo{person}{Zeming Lin}, \bibinfo{person}{Alban Desmaison},
  \bibinfo{person}{Luca Antiga}, {and} \bibinfo{person}{Adam Lerer}.}
  \bibinfo{year}{2017}\natexlab{}.
\newblock \showarticletitle{Automatic differentiation in PyTorch}.
\newblock  (\bibinfo{year}{2017}).
\newblock


\bibitem[\protect\citeauthoryear{Peng, Zhang, Yu, Luo, and Sun}{Peng
  et~al\mbox{.}}{2017}]%
        {peng2017large}
\bibfield{author}{\bibinfo{person}{Chao Peng}, \bibinfo{person}{Xiangyu Zhang},
  \bibinfo{person}{Gang Yu}, \bibinfo{person}{Guiming Luo}, {and}
  \bibinfo{person}{Jian Sun}.} \bibinfo{year}{2017}\natexlab{}.
\newblock \showarticletitle{Large Kernel Matters--Improve Semantic Segmentation
  by Global Convolutional Network}.
\newblock \bibinfo{journal}{\emph{arXiv preprint arXiv:1703.02719}}
  (\bibinfo{year}{2017}).
\newblock


\bibitem[\protect\citeauthoryear{Rhemann, Rother, Wang, Gelautz, Kohli, and
  Rott}{Rhemann et~al\mbox{.}}{2009}]%
        {rhemann2009perceptually}
\bibfield{author}{\bibinfo{person}{Christoph Rhemann}, \bibinfo{person}{Carsten
  Rother}, \bibinfo{person}{Jue Wang}, \bibinfo{person}{Margrit Gelautz},
  \bibinfo{person}{Pushmeet Kohli}, {and} \bibinfo{person}{Pamela Rott}.}
  \bibinfo{year}{2009}\natexlab{}.
\newblock \showarticletitle{A perceptually motivated online benchmark for image
  matting}. In \bibinfo{booktitle}{\emph{Computer Vision and Pattern
  Recognition, 2009. CVPR 2009. IEEE Conference on}}. IEEE,
  \bibinfo{pages}{1826--1833}.
\newblock


\bibitem[\protect\citeauthoryear{Ros, Sellart, Materzynska, Vazquez, and
  Lopez}{Ros et~al\mbox{.}}{2016}]%
        {Ros_2016_CVPR}
\bibfield{author}{\bibinfo{person}{German Ros}, \bibinfo{person}{Laura
  Sellart}, \bibinfo{person}{Joanna Materzynska}, \bibinfo{person}{David
  Vazquez}, {and} \bibinfo{person}{Antonio~M. Lopez}.}
  \bibinfo{year}{2016}\natexlab{}.
\newblock \showarticletitle{The SYNTHIA Dataset: A Large Collection of
  Synthetic Images for Semantic Segmentation of Urban Scenes}. In
  \bibinfo{booktitle}{\emph{The IEEE Conference on Computer Vision and Pattern
  Recognition (CVPR)}}.
\newblock


\bibitem[\protect\citeauthoryear{Shahrian, Rajan, Price, and Cohen}{Shahrian
  et~al\mbox{.}}{2013}]%
        {shahrian2013improving}
\bibfield{author}{\bibinfo{person}{Ehsan Shahrian}, \bibinfo{person}{Deepu
  Rajan}, \bibinfo{person}{Brian Price}, {and} \bibinfo{person}{Scott Cohen}.}
  \bibinfo{year}{2013}\natexlab{}.
\newblock \showarticletitle{Improving image matting using comprehensive
  sampling sets}. In \bibinfo{booktitle}{\emph{Computer Vision and Pattern
  Recognition (CVPR), 2013 IEEE Conference on}}. IEEE,
  \bibinfo{pages}{636--643}.
\newblock


\bibitem[\protect\citeauthoryear{Shen, Tao, Gao, Zhou, and Jia}{Shen
  et~al\mbox{.}}{2016}]%
        {shen2016deep}
\bibfield{author}{\bibinfo{person}{Xiaoyong Shen}, \bibinfo{person}{Xin Tao},
  \bibinfo{person}{Hongyun Gao}, \bibinfo{person}{Chao Zhou}, {and}
  \bibinfo{person}{Jiaya Jia}.} \bibinfo{year}{2016}\natexlab{}.
\newblock \showarticletitle{Deep automatic portrait matting}. In
  \bibinfo{booktitle}{\emph{European Conference on Computer Vision}}. Springer,
  \bibinfo{pages}{92--107}.
\newblock


\bibitem[\protect\citeauthoryear{Sun, Jia, Tang, and Shum}{Sun
  et~al\mbox{.}}{2004}]%
        {sun2004poisson}
\bibfield{author}{\bibinfo{person}{Jian Sun}, \bibinfo{person}{Jiaya Jia},
  \bibinfo{person}{Chi-Keung Tang}, {and} \bibinfo{person}{Heung-Yeung Shum}.}
  \bibinfo{year}{2004}\natexlab{}.
\newblock \showarticletitle{Poisson matting}. In \bibinfo{booktitle}{\emph{ACM
  Transactions on Graphics (ToG)}}, Vol.~\bibinfo{volume}{23}. ACM,
  \bibinfo{pages}{315--321}.
\newblock


\bibitem[\protect\citeauthoryear{Wang and Cohen}{Wang and Cohen}{2007}]%
        {wang2007optimized}
\bibfield{author}{\bibinfo{person}{Jue Wang} {and} \bibinfo{person}{Michael~F
  Cohen}.} \bibinfo{year}{2007}\natexlab{}.
\newblock \showarticletitle{Optimized color sampling for robust matting}. In
  \bibinfo{booktitle}{\emph{Computer Vision and Pattern Recognition, 2007.
  CVPR'07. IEEE Conference on}}. IEEE, \bibinfo{pages}{1--8}.
\newblock


\bibitem[\protect\citeauthoryear{Xu, Price, Cohen, and Huang}{Xu
  et~al\mbox{.}}{2017}]%
        {xu2017deep}
\bibfield{author}{\bibinfo{person}{Ning Xu}, \bibinfo{person}{Brian Price},
  \bibinfo{person}{Scott Cohen}, {and} \bibinfo{person}{Thomas Huang}.}
  \bibinfo{year}{2017}\natexlab{}.
\newblock \showarticletitle{Deep image matting}. In
  \bibinfo{booktitle}{\emph{Computer Vision and Pattern Recognition (CVPR)}}.
\newblock


\bibitem[\protect\citeauthoryear{Yu and Koltun}{Yu and Koltun}{2015}]%
        {yu2015multi}
\bibfield{author}{\bibinfo{person}{Fisher Yu} {and} \bibinfo{person}{Vladlen
  Koltun}.} \bibinfo{year}{2015}\natexlab{}.
\newblock \showarticletitle{Multi-scale context aggregation by dilated
  convolutions}.
\newblock \bibinfo{journal}{\emph{arXiv preprint arXiv:1511.07122}}
  (\bibinfo{year}{2015}).
\newblock


\bibitem[\protect\citeauthoryear{Zhao, Shi, Qi, Wang, and Jia}{Zhao
  et~al\mbox{.}}{2017}]%
        {zhao2017pyramid}
\bibfield{author}{\bibinfo{person}{Hengshuang Zhao}, \bibinfo{person}{Jianping
  Shi}, \bibinfo{person}{Xiaojuan Qi}, \bibinfo{person}{Xiaogang Wang}, {and}
  \bibinfo{person}{Jiaya Jia}.} \bibinfo{year}{2017}\natexlab{}.
\newblock \showarticletitle{Pyramid scene parsing network}. In
  \bibinfo{booktitle}{\emph{IEEE Conf. on Computer Vision and Pattern
  Recognition (CVPR)}}. \bibinfo{pages}{2881--2890}.
\newblock


\bibitem[\protect\citeauthoryear{Zhu, Chen, Wang, Liu, Zhang, and Tang}{Zhu
  et~al\mbox{.}}{2017}]%
        {zhu2017fast}
\bibfield{author}{\bibinfo{person}{Bingke Zhu}, \bibinfo{person}{Yingying
  Chen}, \bibinfo{person}{Jinqiao Wang}, \bibinfo{person}{Si Liu},
  \bibinfo{person}{Bo Zhang}, {and} \bibinfo{person}{Ming Tang}.}
  \bibinfo{year}{2017}\natexlab{}.
\newblock \showarticletitle{Fast Deep Matting for Portrait Animation on Mobile
  Phone}. In \bibinfo{booktitle}{\emph{Proceedings of the 2017 ACM on
  Multimedia Conference}}. ACM, \bibinfo{pages}{297--305}.
\newblock


\end{thebibliography}
